\documentclass[sigconf]{acmart}
\usepackage{subfigure}

\AtBeginDocument{%
  \providecommand\BibTeX{{%
    \normalfont B\kern-0.5em{\scshape i\kern-0.25em b}\kern-0.8em\TeX}}}

\setcopyright{acmcopyright}
\copyrightyear{2021}
\acmYear{2021}
\acmDOI{}

\acmConference[ArXiv]{ArXiv}{December 2019}{cs.AI}
\acmPrice{}
\acmISBN{}

\begin{document}

\title{Node Masking}
\subtitle{Making Graph Neural Networks Generalize and Scale Better}

\author{Pushkar Mishra}
\email{pushkarmishra@fb.com}
\affiliation{%
  \institution{Facebook AI}
  \city{London}
  \country{United Kingdom}
}

\author{Aleksandra Piktus}
\email{piktus@fb.com}
\affiliation{%
  \institution{Facebook AI}
  \city{London}
  \country{United Kingdom}
}

\author{Gerard Goossen}
\email{ggoossen@fb.com}
\affiliation{%
  \institution{Facebook}
  \city{London}
  \country{United Kingdom}
}

\author{Fabrizio Silvestri}
\email{fsilvestri@fb.com}
\affiliation{%
  \institution{Facebook AI}
  \city{London}
  \country{United Kingdom}
}

\renewcommand{\shortauthors}{Mishra et al.}

\begin{abstract}
  \textit{Graph Neural Networks} (\textit{GNNs}) have received a lot of interest in the recent times. From the early \textit{spectral} architectures that could only operate on undirected graphs per a \textit{transductive} learning paradigm to the current state of the art \textit{spatial} ones that can apply \textit{inductively} to arbitrary graphs, GNNs have seen significant contributions from the research community. In this paper, we utilize some theoretical tools to better visualize the operations performed by state of the art spatial GNNs. We analyze the inner workings of these architectures and introduce a simple concept, \textit{Node Masking}, that allows them to generalize and scale better. To empirically validate the concept, we perform several experiments on some widely-used datasets for node classification in both the transductive and inductive settings, hence laying down strong benchmarks for future research.
\end{abstract}



\maketitle

\section{Introduction}
\label{intro}
Graphs are the most effective way of representing different types of entities and relationships amongst them. Several constructs inherently involve the notion of graphs, such as social networks, molecular structures, knowledge bases, recommendation systems, etc. Over the past few year, learning on graphs has become increasingly popular, applications of which can be found in domains ranging from detection of abuse online \cite{mishragcn} and document classification \cite{gcn} to knowledge graph alignment \cite{wang-etal-2018-cross-lingual} and relation extraction in texts \cite{sahu-etal-2019-inter}. Learning on graphs is essentially about leveraging the inductive bias imposed by their relational structures, i.e., \textit{relational inductive bias}, so as to achieve better performance on tasks that can benefit from relation reasoning \cite{DBLP:journals/corr/abs-1806-01261}. The ability to exploit relationships amongst entities in the data is a crucial one for advancing the state of artificial intelligence \cite{tenenbaum2011grow, Lake2017StillNS}.

\begin{figure}[ht]
\centering
\includegraphics[width=7.5cm]{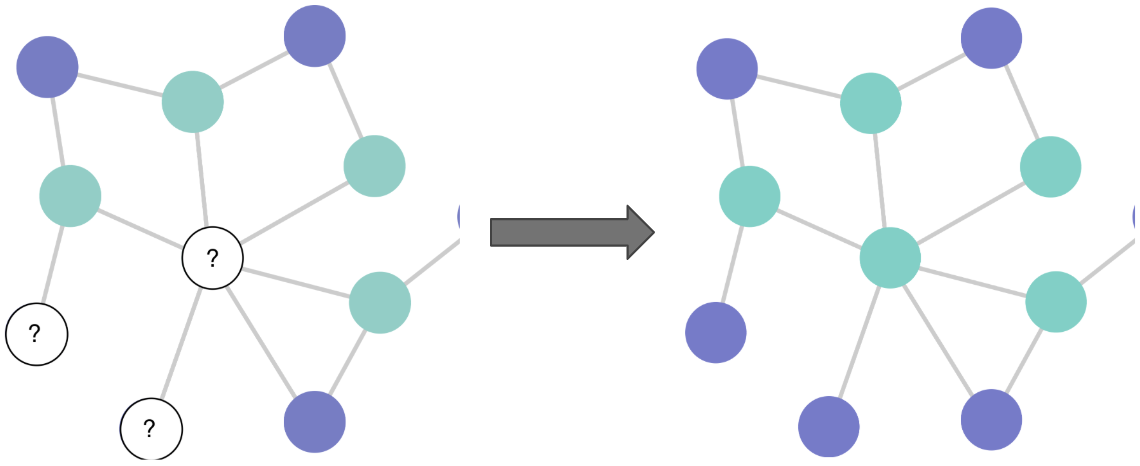}
\caption{Node classification in transductive settings. At the training time, the learning algorithm has access to all the nodes and edges, including those nodes for which labels are to be predicted (denoted by question marks).}
\label{transductive}
\end{figure}

A graph is defined by its set of nodes (i.e., vertices) and its set of edges. There exist two different paradigms for learning on graphs, \textit{transductive} and \textit{inductive}. In transductive learning, the node and edge sets remain constant across the training and prediction phases. In other words, at the training time, the learning algorithm has access to all the nodes and edges, including those for which predictions are to be made. Note that transductive learning does not support generalization to unseen nodes and edges. Figure \ref{transductive} depicts node classification performed in a transductive setting.

\begin{figure}[ht]
    \centering
    \subfigure[A model $\mathcal{H}$ is learned over some graph]{
        \includegraphics[width=6cm]{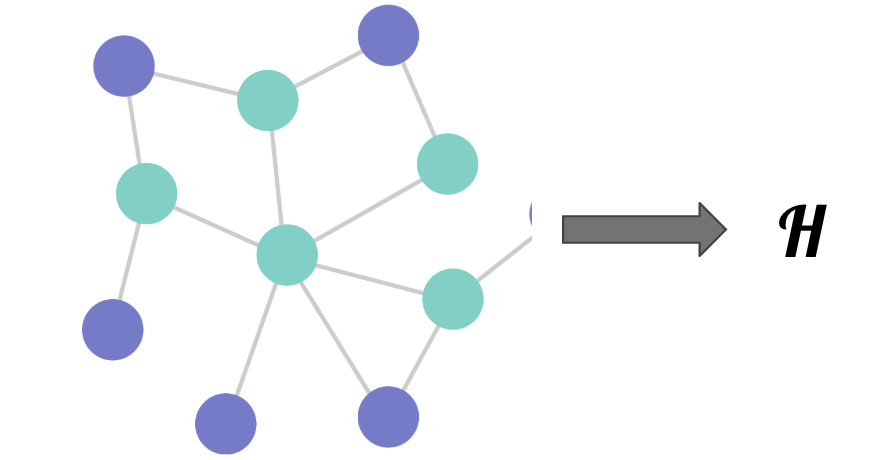}
    }
    \subfigure[The model is then by applied to new nodes and edges]{
        \includegraphics[width=7.5cm]{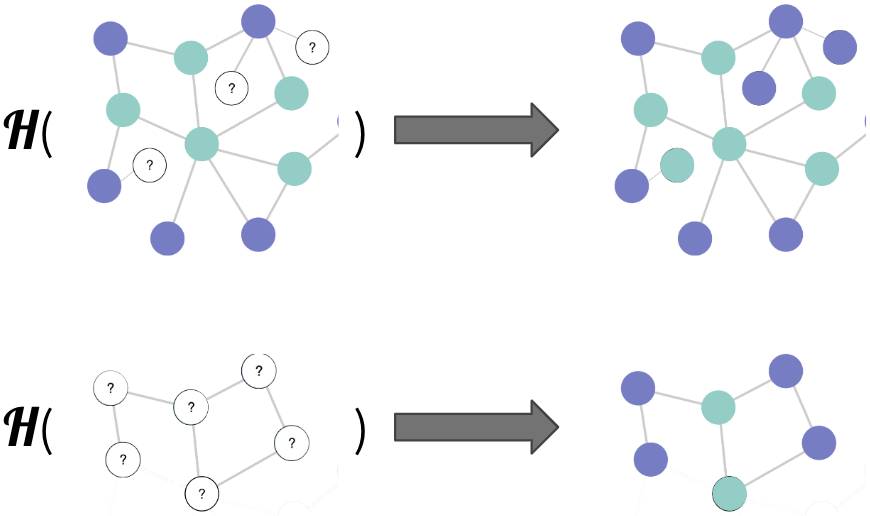}
    }
    \caption{Node classification in inductive settings. Once learned, the model can be applied to new unseen nodes (denoted by question marks). There may or may not exist edges between such new nodes and the nodes used for training.}
    \label{inductive}
\end{figure}

In inductive learning, first a model $\mathcal{H}$ is learned over the \textit{training} graph consisting of some nodes and edges. The learned model is then used to predict on \textit{unseen} nodes and edges that may or may not be disconnected from the nodes and edges in the training graph \cite{clustergcn}. Note that some works \cite{velickovic2018graph,hamilton2017inductive} have instead interpreted inductive learning to mean that the model is first trained on a set of graphs and then applied to a separate set of graphs. But the former interpretation subsumes the latter in that a set of graphs can be treated as a single graph with multiple disconnected components. Figure \ref{inductive} depicts node classification in inductive settings.

Deep learning has brought advancements to several areas within AI. That said, deep learning on graphs is a rather challenging task to perform with traditional architectures like Convolutional Neural Networks or Recurrent Neural Networks. In the recent years, a lot of research has been dedicated to generalizing the convolution operation to graphs \cite{bruna, gcn, hamilton2017inductive, xu2018how}, which has formed the basis of all modern \textit{graph neural networks} (\textit{GNNs}). From the early \textit{spectral} architectures that could only operate on undirected graphs in transductive settings to the current state of the art \textit{spatial} ones that can apply \textit{inductively} to arbitrary graphs, GNNs have undergone significant developments. This paper makes contributions towards further enhancing the capabilities of state of the art spatial GNNs.

\vspace{1mm}
\noindent
\textbf{Our contributions.} We first discuss some theoretical tools to better visualize the operations performed by spatial GNNs. Using these tools, we analyze the inner workings of the state of the art spatial architectures, specifically \textit{aggregation-based} GNNs, to uncover the sources of \textit{over-smoothing} and \textit{over-fitting}. We then propose a simple technique called \textit{Node Masking} to help these GNNs generalize and scale better. We empirically uncover how this technique impacts over-smoothing and over-fitting in GNNs and validate the merits of the technique by performing several experiments on three widely-used benchmark datasets for node classification in both the transductive and inductive settings.

\section{A brief history of GNNs}
\label{history}
The concept of GNNs was first formalized in the work of Gori et al. \cite{gori}. The authors presented GNNs as an extension of \textit{recursive neural networks} whereby they treat nodes as objects denoted by state vectors and edges as the relationships amongst those nodes. Their design consists of two main steps: i) iterative update of nodes' state vectors based on the labels and state vectors of their neighbors up to a stable fixed point, and ii) back-propagation for adjustment of parameters used in the update step. This approach was further refined in the work of Scarselli et al. \cite{scarselli}.

\subsection{Spectral GNNs}
Bruna et al. \cite{bruna} laid the foundation for generalization of the \textit{convolution} operation from regular grids to graphs. Leveraging \textit{spectral graph theory}, they proposed an architecture for performing \textit{spectral convolutions} on graphs. Given a graph $G$, their architecture considers the feature vectors on nodes as multi-channel graph signals. It learns \textit{spectral filters} that act on these signals in the \textit{Fourier} domain defined by the eigenvectors of the \textit{Laplacian} matrix of $G$. This architecture displays limited scalability since the filters learned are not localized, i.e., they act on the whole graph, and computation of the Laplacian's eigenvectors is itself an expensive operation.

To overcome these issues, Defferrard et al. \cite{defferrard2016convolutional} and Levie et al. \cite{levie2017cayleynets} proposed spectral architectures, \textit{ChebNet} and \textit{CayleyNet} respectively, comprising localized filters approximated by \textit{Chebyshev} and \textit{Cayley} polynomials. Kipf and Welling \cite{gcn} further simplified ChebNet by making the filters localized to 1-hop neighborhoods. By stacking multiple such filters in layers, they showed that any number of hops could be covered in the convolution operation. They called this new architecture \textit{Graph Convolutional Network} (\textit{GCN}). Note that all spectral architectures are inherently transductive in that the filters learned on a graph are specific to the eigenbasis of its Laplacian. This not only limits the ability of these architectures to generalize to new nodes and edges. Although GCN itself is spectral, the idea of stacking multiple layers to cover higher-order neighborhoods led to the concept of spatial GNNs.

\subsection{Spatial GNNs}
Spatial GNNs define the convolution operation directly on the structure of the graph. In other words, they work by learning functions to compute representations for nodes or edges that capture the features and structures of their surrounding neighborhoods. Once learned, these functions can then be inductively applied to new nodes and edges. Spatial GNNs are preferred over their spectral counterparts due to their scalability, inductive capabilities, and ability to handle myriad types of graphs \cite{gnn-survey}.

\subsubsection{Sampling-based}
\textit{GraphSAGE} \cite{hamilton2017inductive} was one of the first spatial GNNs. For a node $v$, \textit{GraphSAGE} randomly samples a fixed number of nodes from its $K$-hop neighborhood and learns to computes a representation for $v$ based on its own features plus the features of its sampled neighbors. Note that this design does not exhibit \textit{structural invariance}, i.e., the GNN cannot accommodate varying neighborhood structures but rather constrains nodes to a fixed number of neighbors only. A clear drawback of such a sampling-based design is that a lot of neighborhood information is wasted.

\subsubsection{Aggregation-based}
Aggregation-based spatial GNNs eliminate the need for sampling fixed number of neighbors. They work by iteratively computing representations for nodes based on those of their respective neighbors. A $k$-layer aggregation-based GNN sequentially performs such updates $k$ times, consequently computing a representation for every node that captures its $k$-hop neighborhood. The update operation performed by the $k^{th}$ layer for a node $v$ can be stated as:
\begin{equation}
\label{equation1}
    h^{(k)}_v = f^{(k)}\,\left(\,h^{(k-1)}_v,\,g\,(\{h^{(k-1)}_u: u\in\mathcal{N}_v\})\,\right)
\end{equation}

\noindent
where $\mathcal{N}_v$ denotes the set of neighbors of $v$, and $h^{(0)}_v$ is the input feature vector of $v$. The \textit{aggregate} function $g$ aggregates representations of neighbors, and the \textit{combine} function $f$ combines the aggregated representation with that of $v$ itself. This formulation is general enough to cover the various aggregation-based GNNs that exist. All of them mainly differ in their choice of $f$ and $g$. Note that the compute required to capture $k$-hop neighborhood grows linearly with $k$ in aggregation-based GNNs as opposed to exponentially in sampling-based GNNs.

Velickovic et al. \cite{velickovic2018graph} proposed \textit{Graph Attention Networks} (\textit{GAT}) wherein a node's representation is iteratively updated by aggregating the representations of its neighbors combining them with that of the node's as per coefficients allocated by a \textit{self-attention} mechanism. The defines the update operation in the $k^{th}$ layer as:
\begin{equation}
\label{equation2}
    h^{(k)}_v = \bigg\Vert_l\,\sigma\left(\,\sum_u^{v\,\cup\,\mathcal{N}_v}\alpha^{(k)}_l(v, u)\cdot W^{(k)}_lh^{(k-1)}_u\,\right)
\end{equation}
\noindent
where $\alpha^{(k)}_l(v, u)$ is the attention coefficient of node $u$ with respect to node $v$ from the $l^{th}$ attention head, $W_l$ is the weight matrix for the $l^{th}$ attention head, and $\Vert$ denotes concatenation over the heads.

Xu et al. \cite{xu2018how} recently introduced the \textit{Graph Isomorphism Network} (\textit{GIN}) whose theoretical foundations allows it to be maximally powerful amongst the various spatial GNNs. The GIN-$\epsilon$ architecture defines the update operation as:
\begin{equation}
\label{equation3}
    h^{(k)}_v = MLP^{(k)}\left((1 + \epsilon^{(k)})\cdot h^{(k-1)}_v + \sum_u^{\mathcal{N}_v}h^{(k-1)}_u\right)
\end{equation}

\noindent
where $MLP$ represents a multi-layer perceptron.

In this paper, we focus on aggregation-based GNNs given that they yield state of the art \cite{gnn-survey} performance. We work with the GAT and GIN architectures only but our techniques can be applied to other aggregation-based GNNs.


\section{Theoretical Framework}
\label{theory}
Hereon, we assume that the graphs we consider are undirected, implying that an edge can be traversed from either endpoints. That said, the work presented in this paper is trivially applicable to directed graphs too. We also assume that all nodes within a given graph are uniquely identifiable.

\subsection{Aggregation trees}
We discuss the concept of \textit{aggregation trees} as the theoretical tool for visualizing aggregation-based GNNs. Aggregation trees have been explored in the past as graph kernels \cite{xu2018how} under names like \textit{tree-walks} \cite{tree-walks} or \textit{subtree patterns} \cite{subtree-patterns}.

\begin{figure*}[ht]
    \centering
    \subfigure[An undirected graph]{
        \includegraphics[width=3.25cm]{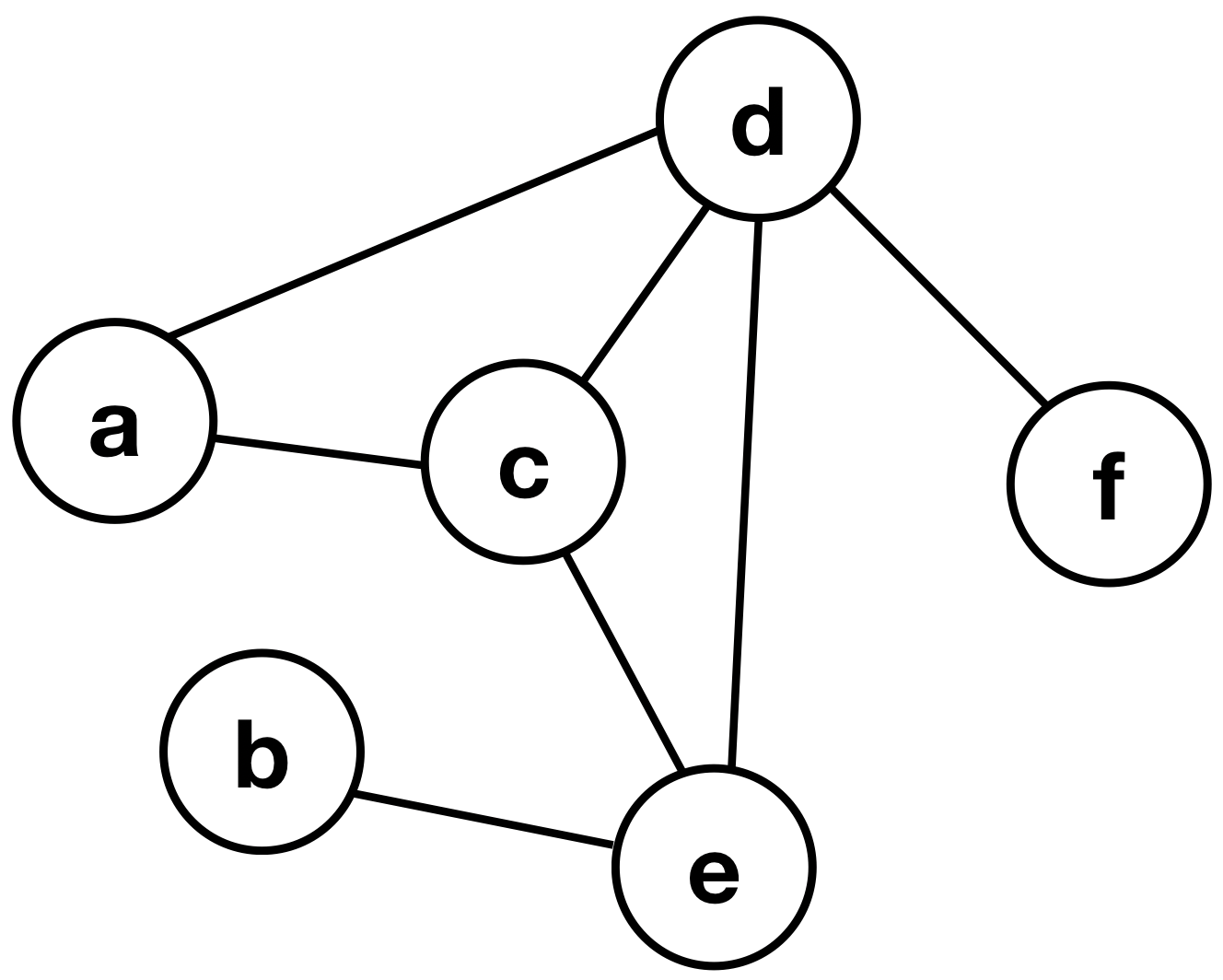}
    }
    \qquad\quad
    \subfigure[$T_a^0$]{
        \hspace{1mm}\includegraphics[width=0.65cm]{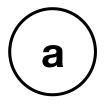}\hspace{1mm}
    }
    \qquad\quad
    \subfigure[$T_a^1$]{
        \includegraphics[width=3.5cm]{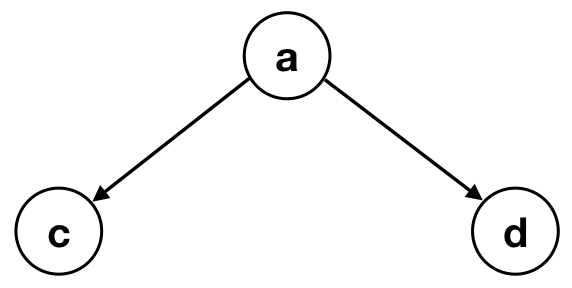}
    }
    \qquad\quad
    \subfigure[$T_a^2$]{
        \includegraphics[width=5.25cm]{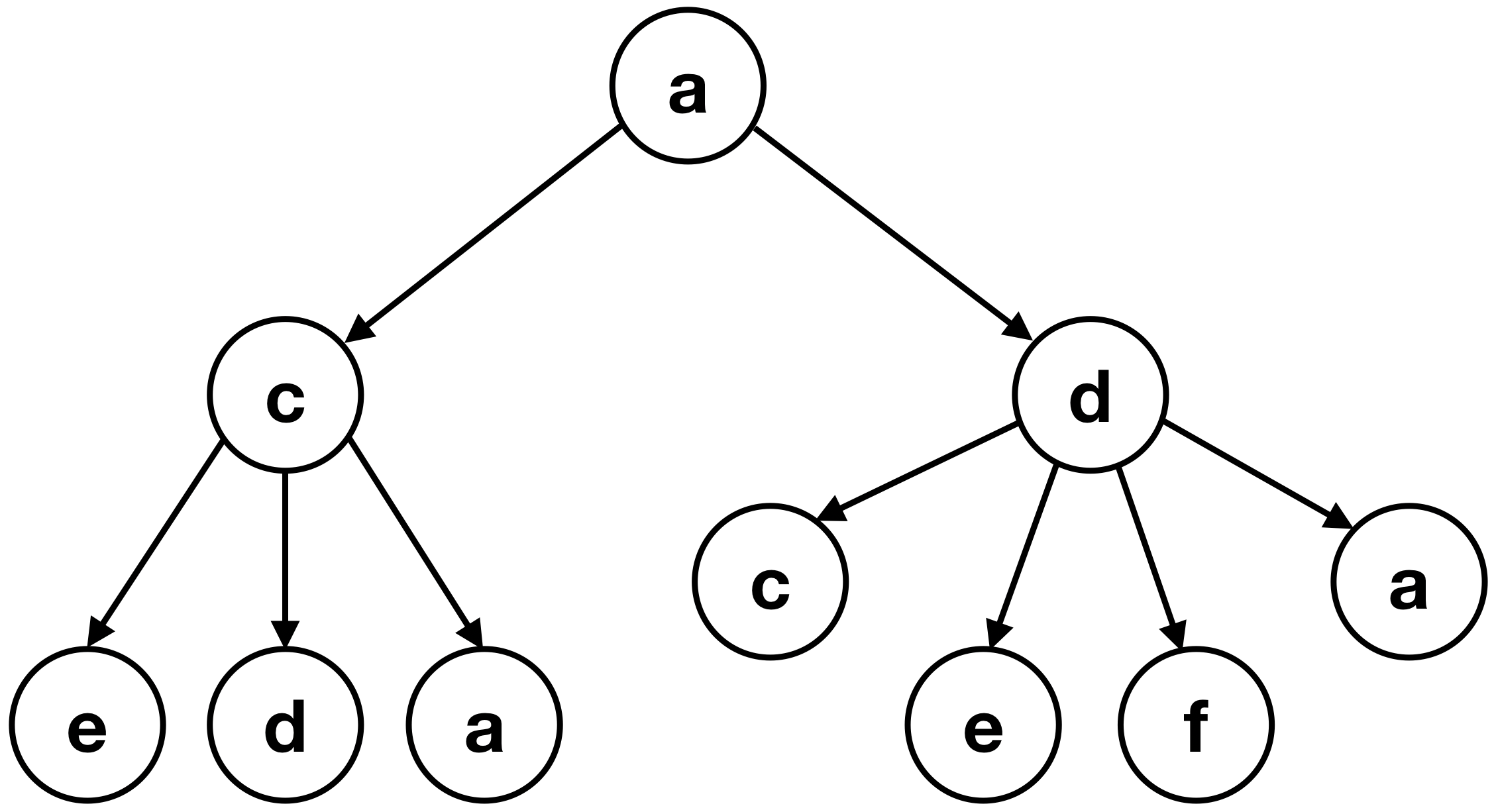}
    }
    \caption{Sub-figure (a) shows an undirected graph, i.e., edges can be walked in either directions. Sub-figures (b-d) show some aggregation trees $T^k_a$ of node \textit{a}. Every path from the root of a $T^k_a$ to some leaf in it is a valid $k$-hop walk in the graph starting at node $a$, and vice versa.}
    \label{graph-tree}
\end{figure*}

\begin{definition}
Given a graph $G$ with a node $v$ in it, let $\mathcal{P}_v^k$ denote the set of all possible \textit{walks} of length $k$, i.e., of $k$ hops, starting at $v$. Walks are paths with possibly repeated nodes. The $k$-\textit{aggregation tree} $T_v^k$ of a node $v$ is the smallest \textit{arborescence} \cite{arbor} rooted at $v$ such that $w$ is a path from the root of $T_v^k$ to a leaf in it if and only if $w \in \mathcal{P}_v^k$. Here, smallness is by the number of nodes. Note that $T^k_v$ is of height $k$ with all the leaves at the same depth. We refer to $G$ itself as the \textit{base} graph of $T^k_v$.
\end{definition}

\begin{figure}[ht]
    \centering
    \subfigure[Not a valid arborescence]{
        \includegraphics[width=3.45cm]{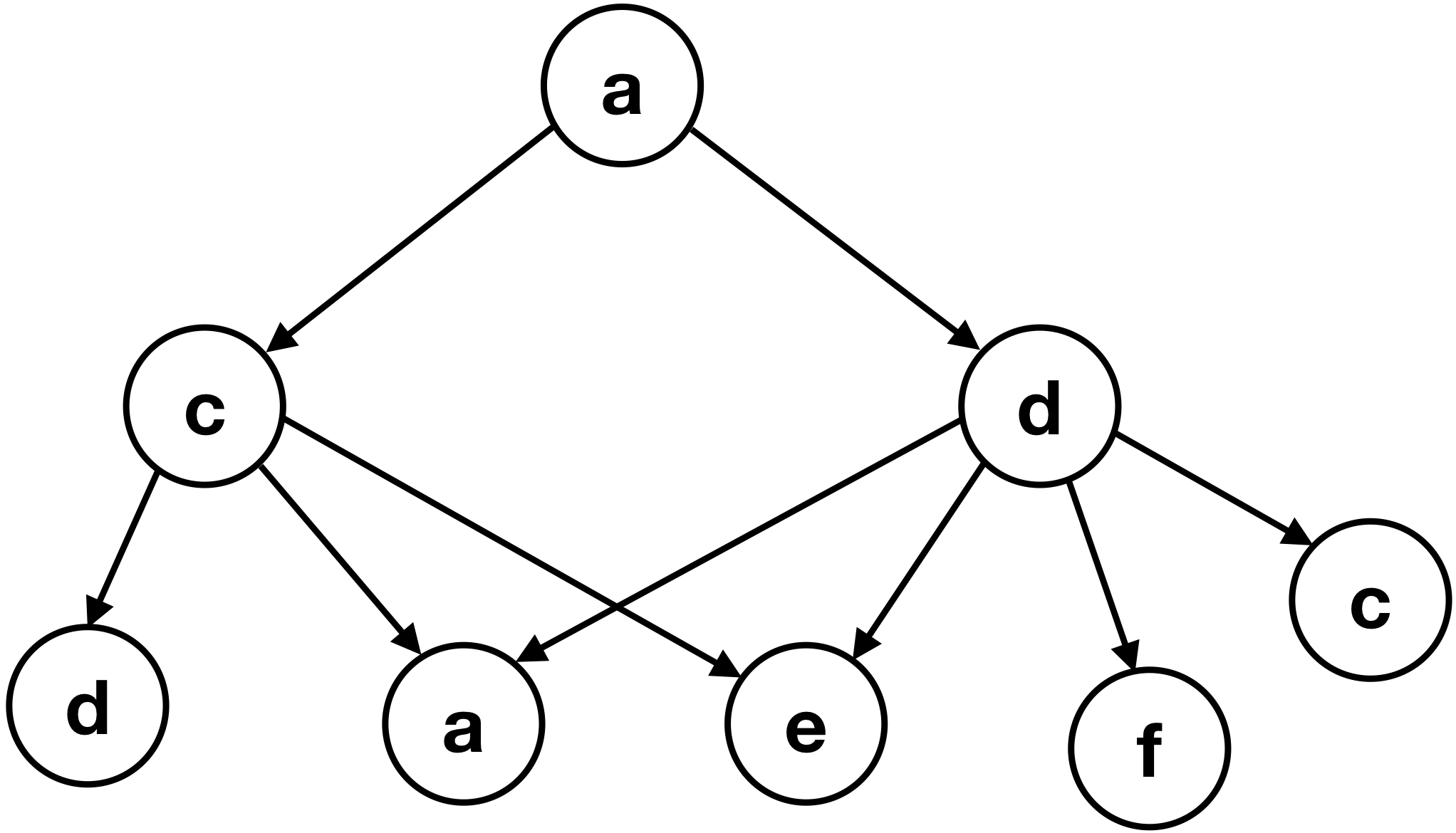}
    }
    \qquad
    \subfigure[Not the smallest possible]{
        \includegraphics[width=3.65cm]{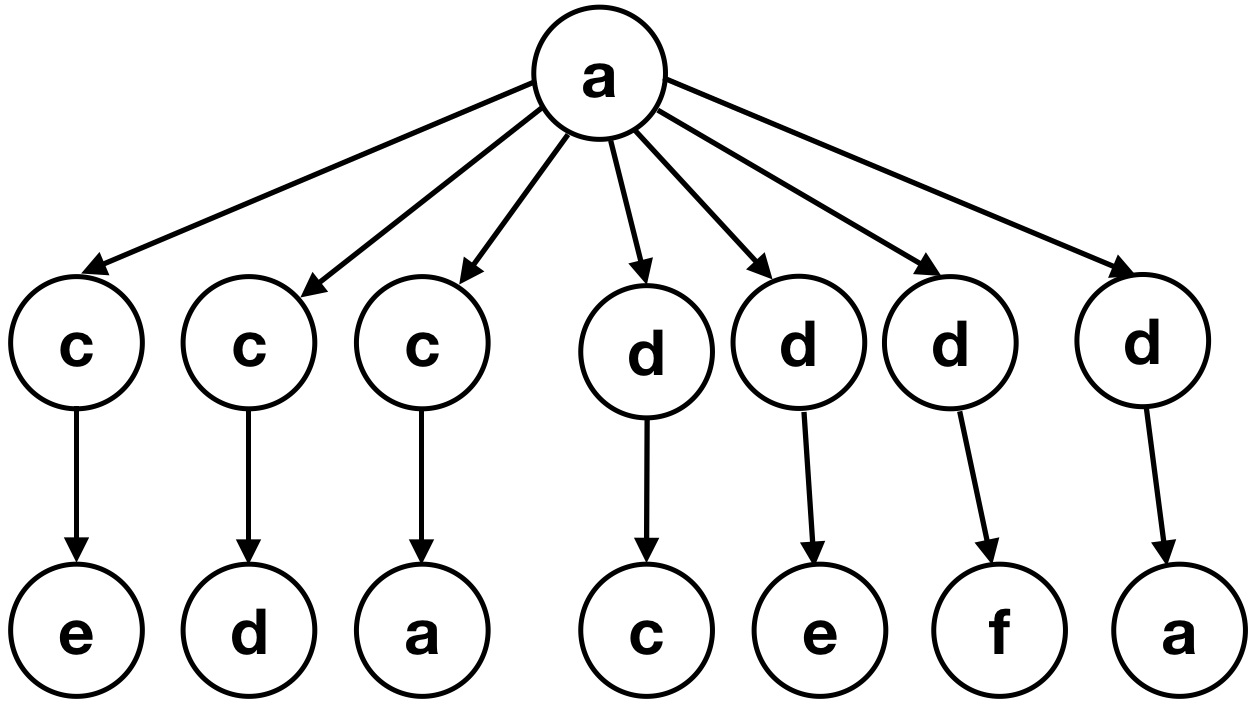}
    }
    \caption{Trees having the required paths from the root to leaves, but not the valid 2-aggregation tree of $a$. Sub-figure (a) is not a valid arborescence since there should only be one path from the root to any node in an arborescence \cite{gordon1989greedoid}. Sub-figure (b) is a valid arborescence but not the smallest.}
    \label{non-tree}
\end{figure}

Figure \ref{graph-tree} shows an example of a graph along with the various aggregation trees $T^*_a$ of the node $a$ in it. Figure \ref{non-tree} shows some trees that have the same root node and the same set of paths from the root to leaves but are not valid $2$-aggregation trees of node $a$. We would like to point out to the reader that the two assumptions we made for graphs are not upheld in the case of aggregation trees:

\begin{itemize}
\item Aggregation trees can contain multiple replicas of the same node \cite{subtree-patterns}. For example, the $2$-aggregation tree $T^2_a$ shown in figure \ref{graph-tree} has node $a$ as the root and also as two of the leaves. The replicas are treated as distinct nodes so that aggregation trees remain acyclic, but all of them correspond to the same node in the base graph.
\item Aggregation trees are directed, making the notion of neighbor set different in their case than in the case of undirected graphs. In an aggregation tree, the neighbor set $\mathcal{N}_v$ of a node $v$ only contains those nodes that have incoming edges from $v$, but not those that have outgoing edges to $v$. Therefore, in the context of aggregation trees, we refer to the neighbor set of a node as its \textit{set of children} or \textit{child set} for clarity.
\end{itemize}

The two lemmas given below highlight some core properties of aggregation trees.
\begin{lemma}
\label{lemma-subtree}
Given a $k$-aggregation tree $T^k_v$, let a subtree of $T^k_v$ be defined as any node along with all its descendants up to some depth $l$ ($l \leq k$). Then every subtree of $T^k_v$ is also an aggregation tree.
\end{lemma}

\textit{Proof.} Let us assume that the subtree at some node $u$ in $T^k_v$ is not an aggregation tree. Then there exists a path in $T^k_v$ from the root $v$ to some leaf in it via $u$ that is not a $k$-length walk starting at node $v$ in the base graph, or vice versa. But this contradicts the definition of aggregation trees. Hence, such a subtree cannot exist. \hfill$\square$

\begin{lemma}
\label{lemma-neighbor}
Given a graph $G$, let $v$ be any node in it without loss of generality. The child set of any non-leaf node in $T^k_v$ is equal to the neighbor set of the corresponding node in $G$.
\end{lemma}

\textit{Proof.} We prove this also by contradiction. A non-leaf node in $T^k_v$ must be reachable in at most $k-1$ hops from $v$ in $G$. Consequently, if there exists a non-leaf node in $T^k_v$ such that its child set is not equal to its corresponding neighbor set in $G$, then there exists a path from the root of $T^k_v$ to a leaf in it that is not a $K$-length walk in $G$, or vice versa. However, this contradicts the definition of $k$-aggregation trees. Hence, such a non-leaf node cannot exist.\hfill$\square$

\subsection{Connection to aggregation-based GNNs}
We now establish the connection between aggregation trees and aggregation-based GNNs.

\begin{theorem}
\label{k-tree-theorem}
Given a graph $G$, let $v$ be any node in it without loss of generality. The $k$-aggregation tree $T^k_v$ of $v$ denotes the structure captured by the representation $h^{(k)}_v$ computed using a $k$-layer aggregation-based GNN for $v$. Alternatively, a $k$-layer aggregation-based GNN, when applied to $T^k_v$, computes the same representation $h^{(k)}_v$ for root $v$ as it does for node $v$ when applied to $G$.
\end{theorem}

\textit{Proof.} We prove this by induction on $k$. As per equation \ref{equation1}, when $k = 1$, the computation of $h^{(1)}_v$ by a $1$-layer aggregation-based GNN is given by:
\begin{equation*}
    h^{(1)}_v = f^{(1)}\,\left(\,h^{(0)}_v,\,g\,(\{h^{(0)}_u: u\in\mathcal{N}_v\})\,\right)
\end{equation*}
By lemma \ref{lemma-neighbor}, we have that the child set of root $v$ in $T^1_v$ is identical to the neighbor set of $v$ in $G$. Furthermore, $h^{(0)}_*$ are the input feature vectors for nodes; any node in $T^1_v$ has the same $h^{(0)}_*$ as its corresponding node in $G$. So, the theorem trivially holds for $k=1$. Assume that the theorem also holds for some $k > 1$, i.e., $h^{(k)}_v$ computed for root $v$ of $T^k_v$ by a $k$-layer aggregation-based GNN is identical to $h^{(k)}_v$ computed for node $v$ in $G$. Now, the computation of $h^{(k + 1)}_v$ by a $k+1$-layer aggregation-based GNN is given by:
\begin{equation*}
    h^{(k+1)}_v = f^{(k+1)}\,\left(\,h^{(k)}_v,\,g\,(\{h^{(k)}_u: u\in\mathcal{N}_v\})\,\right)
\end{equation*}
As before, we have that the child set of root $v$ in $T^{k+1}_v$ is identical to the neighbor set of $v$ in $G$. Additionally, the representations $h^{(k)}_*$ for root $v$ and its children capture the respective subtrees of depth $k$ under them. Since, these subtrees are actually $k$-aggregation trees (lemma \ref{lemma-subtree}), following our assumption, $h^{(k)}_*$ must be the same for root $v$ and its children as for the corresponding nodes in $G$. So, the theorem holds for $k+1$ when it holds for $k$ because inputs to the computation of $h^{(k+1)}_v$ are the same in the case of $T^{k+1}_v$ and $G$. Hence, the theorem holds for $k >= 1$.\hfill$\square$

Essentially, the $k$-aggregation tree $T^k_v$ is a visual depiction of the representation $h^{(k)}_v$ computed by a $k$-layer aggregation-based GNN. Every subtree in $T^k_v$ is the depiction of some representation $h^{(l)}_*$ ($l < k$) computed intermediately.


\section{Analysis of aggregation-based GNNs}
\label{analysis}
Aggregation trees surface two important issues stemming from the way that aggregation-based GNNs operate.

The first one concerns the \textit{generalization} ability of these architectures. When computing the representation $h^{(k)}_v$, a $k$-layer aggregation-based GNN with $k > 1$ aggregates from the same set of nodes multiple times, especially from $v$ itself. An example of this can be seen in figure \ref{graph-tree} where the node $a$ appears multiple times in its own $2$-aggregation tree $T^2_a$. Such repetitions allow the GNN to amplify the interactions (i.e., the message passing) between nodes and their neighbors, which has two known downsides \cite{dropedge}: 1) it facilitates over-smoothing whereby representations of nodes within a neighborhood become indistinguishable despite belonging to different classes, and 2) it results in over-fitting whereby the GNN cannot generalize well beyond training. In fact, Wang et al. \cite{wang2019improving} even present empirical observations of over-smoothing in GAT.

\begin{figure}[ht]
    \centering
    \subfigure[Node $v$ enters the graph post training]{
        \hspace{2cm}\includegraphics[width=3.5cm]{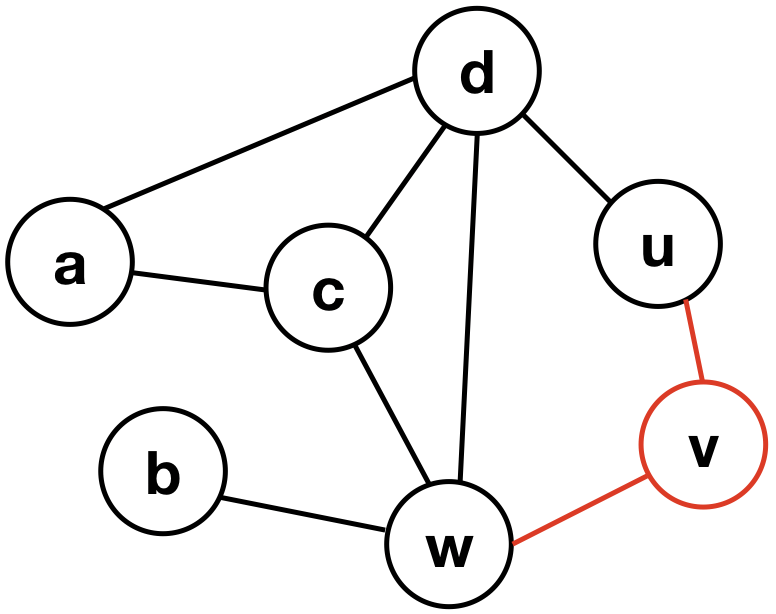}\hspace{2cm}
    }
    \subfigure[$T^k_v$, the $k$-aggregation tree of $v$]{
        \includegraphics[width=5.75cm]{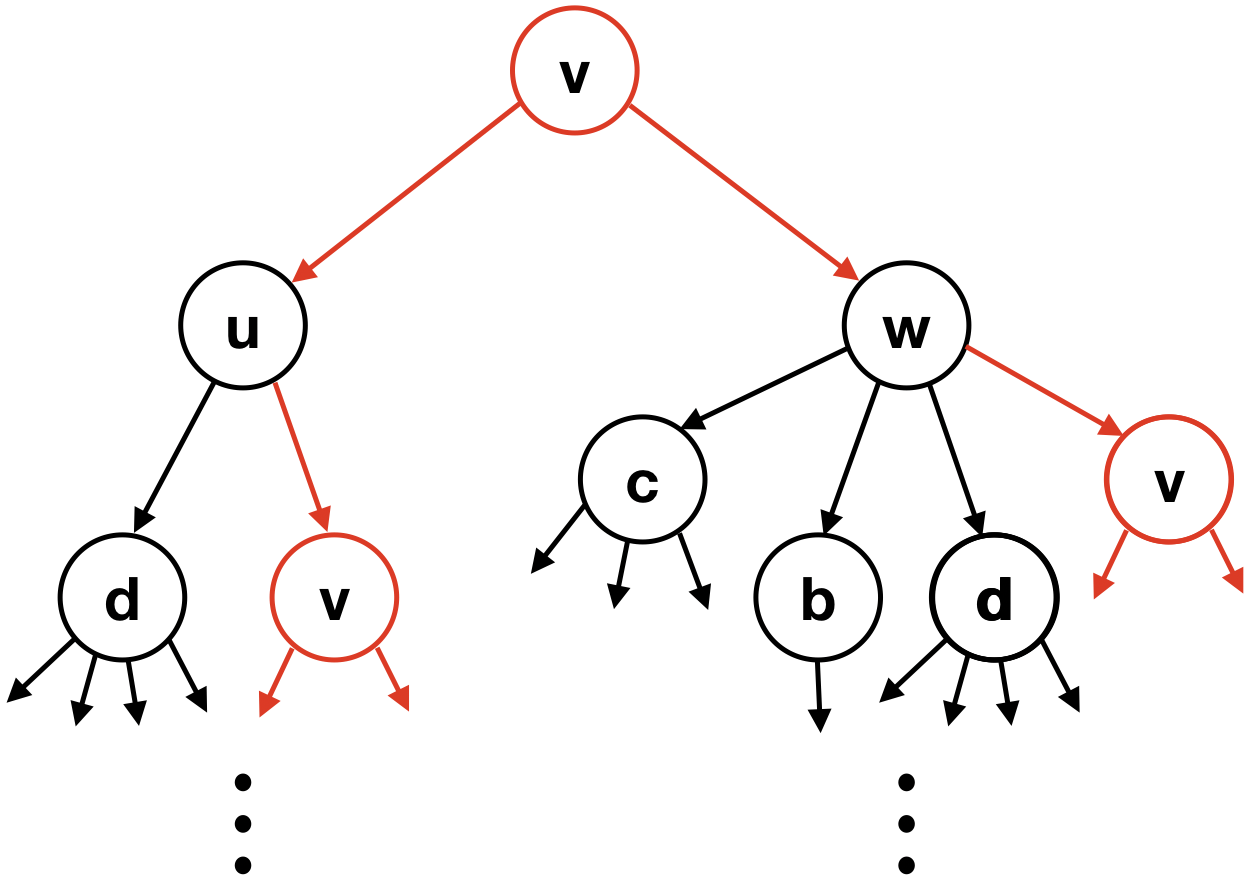}
    }
    \caption{Depiction of the scenario where a node $v$ enters a graph at prediction time after training has finished. The $k$-aggregation tree of node $v$ is also shown.}
    \label{ind-no-cache}
\end{figure}

The second issue pertains to \textit{scalability} of a $k$-layer aggregation-based GNN where $k > 1$. In many real-world settings, there is a large graph that keeps growing with time. A prime example is social networks where new users join from time to time. In such settings, learning follows the inductive paradigm whereby we train the GNN on a snapshot of the graph in time, and then predict on new nodes that enter the graph after that. Figure \ref{ind-no-cache}(a) depicts such a setting where $v$ is a node entering the graph after training. Figure \ref{ind-no-cache}(b) shows the $k$-aggregation tree $T^k_v$, i.e., the structure that will be captured by the representation $h^{(k)}_v$. Essentially, the GNN requires input vectors $h^{(0)}_*$ for all the nodes within $k$ hops of $v$ to compute $h^{(k)}_v$. This can be up to $\mathcal{O}(d^k)$ representations in total, where $d$ is the average degree of a node.

\begin{figure}[ht]
    \centering
    \includegraphics[width=5.5cm]{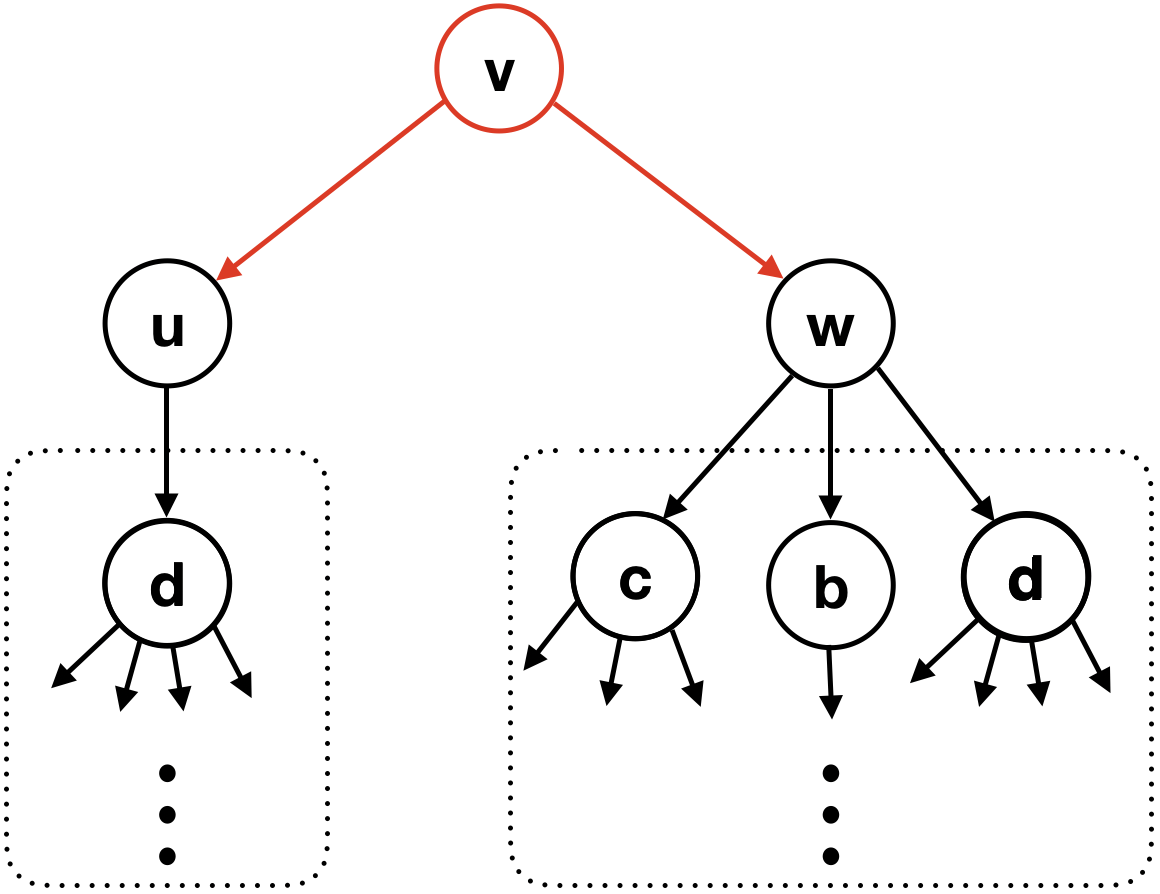}
    \caption{The $k$-aggregation tree of node $v$ as per equation \ref{equation4}. The dotted boxes encapsulate the aggregation trees captured by cached representations $\Tilde{h}^{(1)}_*$, $\dots$, $\Tilde{h}^{(k-1)}_*$ for $u$ and $w$.}
    \label{ind-cache}
\end{figure}

Storing representations from layers of a GNN has been explored as an optimization before \cite{chen2018stochastic}. In the example above, the representations $h^{(*)}_u$ and $h^{(*)}_w$ can be \textit{cached} at the end of training since they capture the relevant aggregation trees of $u$ and $w$ respectively. Then the representation $\hat{h}^{(k)}_v$ can be approximated at prediction time as:
\begin{equation}
\label{equation4}
    \hat{h}^{(k)}_v = f^{(k)}\,\left(\,\hat{h}^{(k-1)}_v,\,g\,(\{\Tilde{h}^{(k-1)}_u: u\in\mathcal{N}_v\})\,\right)
\end{equation}
where $\Tilde{h}$ denotes the cached representations, and $\hat{h}$ signifies that the computed representation is an approximation of $h^{(k)}_v$. The $k$-aggregation tree for equation \ref{equation4} is shown in figure \ref{ind-cache}. Note that, unlike \ref{ind-no-cache}(b), now $v$ only appears as the root, making $h^{(k)}_v$ and $\hat{h}^{(k)}_v$ different. This is because $v$ was not present during training, and consequently, not covered by the cached representations.

Essentially, with caching, the GNN only requires the $k$ representations $\Tilde{h}^{(0)}_*$, $\dots$, $\Tilde{h}^{(k-1)}_*$ for every 1-hop neighbor of $v$ to approximate $h^{(k)}_v$, i.e., up to $\mathcal{O}(kd)$ representations in total. While this is a substantial gain in efficiency, we show empirically in section~\ref{experiments} that the performance at prediction suffers given that the structure of aggregation trees differs between training and prediction times.


\section{Node Masking}
\label{node-mask}
We propose \textit{node masking} as a novel yet simple training phase technique to address the issues highlighted in the previous section.

\subsection{Description}
We begin by formalizing the notion of a \textit{masking function} for generic countable sets.

\begin{definition}
Let $\mathcal{S}$ be a set of elements; we assume $\mathcal{S}$ is countable. Let $\mathcal{B}^n_p$ be the set of outputs of $n$ Bernoulli trials ($n = |\mathcal{S}|)$ with probability $p$ of success. We define $\delta_p$ to be a bijective mapping from $\mathcal{S}$ to $\mathcal{B}^n_p$, and refer to it as the \textit{Bernoulli select function}. There can be $n!$ different $\delta_p$. Now, the \textit{masking function} $\mathcal{M}$ can be defined for $\mathcal{S}$ and a $\delta_p$ as:
\begin{equation}
\label{equation5}
    \mathcal{M}(\mathcal{S}, \delta_p) = \{e : e \in \mathcal{S},\,\delta_p(e) = 1\}
\end{equation}
\end{definition}

Here, all the elements $e$ with $\delta_p(e) = 0$ are said to be \textit{masked}. Next, we demonstrate how we inculcate this masking function $\mathcal{M}$ in the computations of aggregation-based GNNs to tackle the issues we discussed in the previous section. Let $\delta^G_p$ be the Bernoulli select function over the set of nodes of a given graph $G$. We propose \textit{node masking} as the following modification to equation \ref{equation1} that defines the update operation in the $k^{th}$ layer of an aggregation-based GNN:
\begin{equation}
\label{equation6}
    h^{(k)}_v = f^{(k)}\bigg(h^{(k-1)}_v,g\Big(\{h^{(k-1)}_u: u\in\mathcal{M}(\mathcal{N}_v,\delta^G_p)\}\Big)\bigg)
\end{equation}

We refer to $p$ as the \textit{node masking rate}. If $p$ is set to 1 in $\delta^G_p$, then equation \ref{equation6} resembles equation \ref{equation1}, i.e., node masking has no effect. Thus, we say node masking is \textit{inactive} when $p = 1$ and \textit{active} when $p < 1$.

The actual implementation of node masking may vary across the different aggregation-based GNNs. Equations \ref{equation7} and \ref{equation8} lay out the implementations of node masking in GAT and GIN-0 \cite{xu2018how} architectures respectively:

\begin{equation}
\label{equation7}
    h^{(k)}_v = \bigg\Vert_l\,\sigma\left(\,\sum_u^{v\,\cup\,\mathcal{N}_v}\delta^G_p(u)\cdot\alpha^{(k)}_l(v, u)\cdot W^{(k)}_lh^{(k-1)}_u\,\right)
\end{equation}
\begin{equation}
\label{equation8}
    h^{(k)}_v = MLP^{(k)}\,\left(h^{(k-1)}_v + \,\sum_u^{\mathcal{N}_v}\,\delta^G_p(u)\cdot h^{(k-1)}_u\,\right)
\end{equation}

In both cases, any node $u$ with $\delta^G_p(u) = 0$ is effectively masked since its contribution to the sum is nullified. 

\begin{figure}[ht]
    \centering
    \subfigure[No nodes masked]{
        \includegraphics[width=4.5cm]{sample-tree.png}
    }
    \subfigure[Node $d$ is masked]{
        \includegraphics[width=2.5cm]{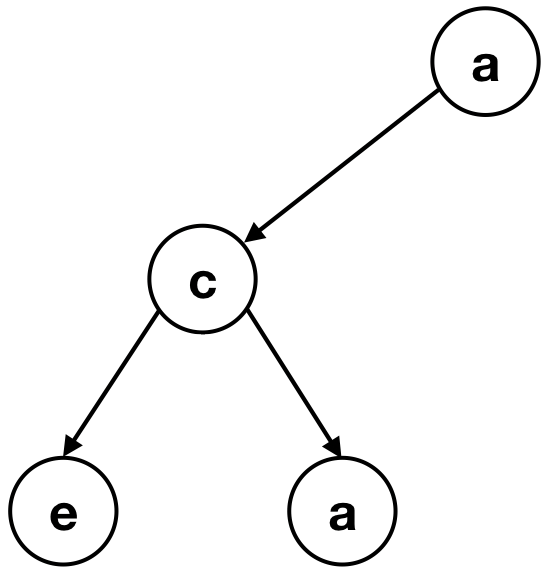}
    }
    \subfigure[Node $a$ is masked]{
        \includegraphics[width=3.75cm]{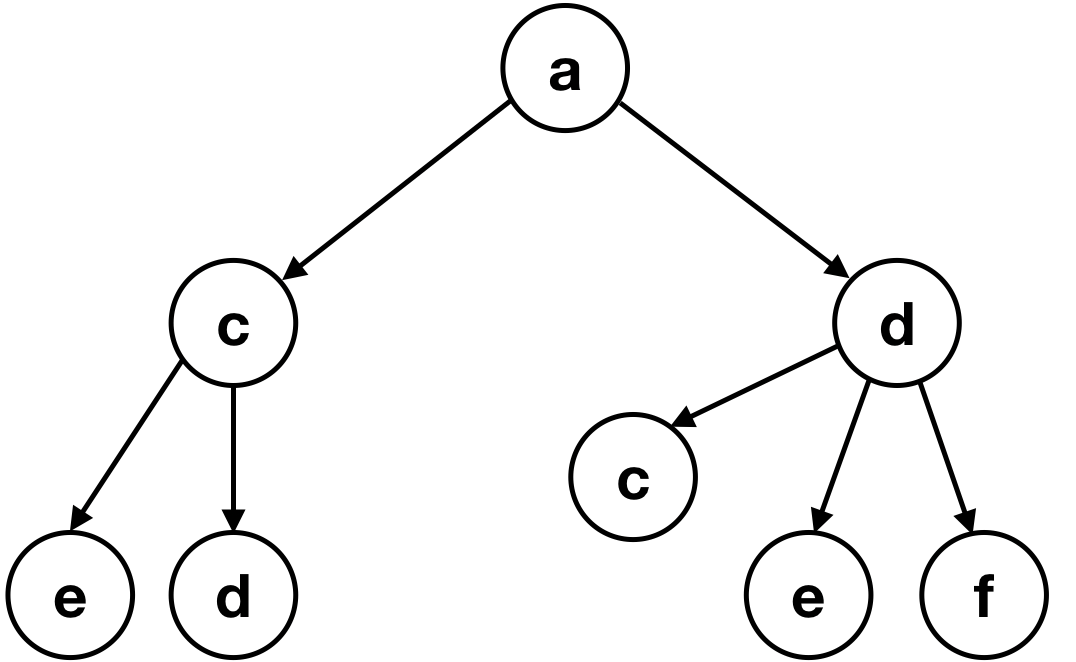}
    }
    \subfigure[Nodes $a$ and $f$ are masked]{
        \includegraphics[width=4cm]{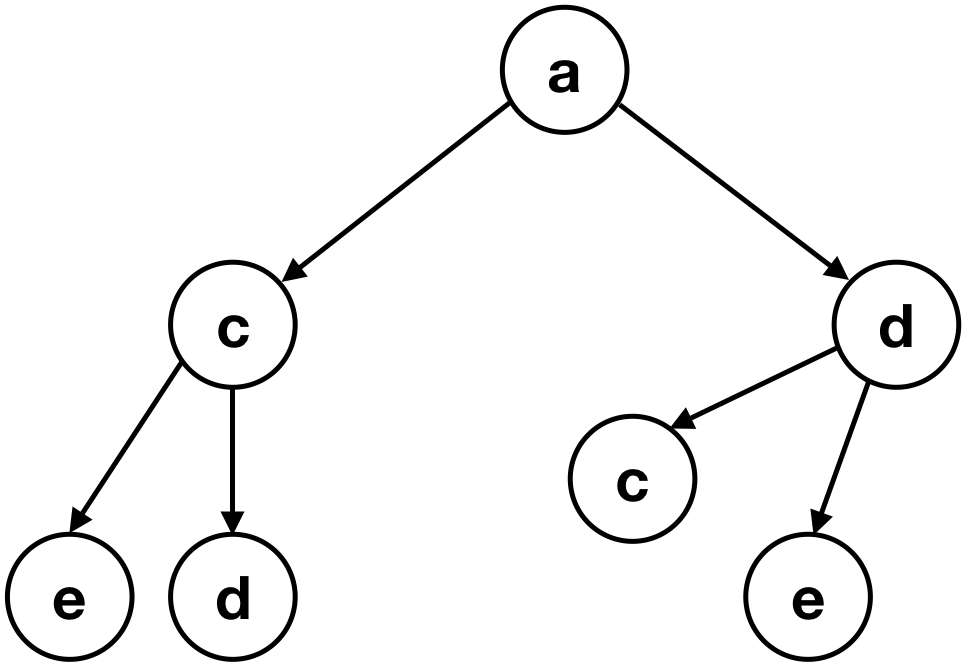}
    }
    \caption{Some possible $2$-aggregation trees $T^2_a$ of node $a$ in the base graph shown by figure \ref{graph-tree}(a). The aggregation trees vary based on the nodes that are masked. If a node is masked, it is excluded from the child set of other nodes but can still appear as the root.}
    \label{masked-trees}
\end{figure}

Essentially, given a graph $G$ and a $k$-layer aggregation-based GNN, if a node $v$ in $G$ is masked, then the representations $h^{(l)}_v$ ($l < k$) are discarded during the computations performed by the GNN. The equivalent effect in aggregation trees is that $v$ is is not included in the child set of other nodes in any aggregation tree since it cannot contribute to representations of other nodes. Going back to the example graph in figure \ref{graph-tree}, figure \ref{masked-trees} shows some possible $2$-aggregation trees $T^2_a$ of the node $a$ depending on the nodes that are masked. Note, even if a node is masked, it can still appear as the root of its own aggregation tree since root is not in any child set.

When a $k$-layer aggregation-based GNN with active node masking is trained on some graph $G$, a Bernoulli select function $\delta^G_p$ is randomly sampled in every training epoch, allowing the GNN to see many different $T^k_v$ for every node $v$. This has two advantages. First, the GNN is discouraged from simply associating a node and its neighbors together. Second, if $\delta^G_p(v)=0$ in an epoch, then $T^k_v$ has the same structure as the aggregation tree in figure \ref{ind-cache}, i.e., no repetition of $v$ in $T^k_v$. This reduces the tendency of the GNN to focus heavily on $v$'s own features and also sensitizes the GNN to that structure of the aggregation trees that it may encounter at prediction time if caching is used.

In summary, node masking can be seen a training phase technique for GNNs that stresses the relational inductive biases in the data, while also having a regularizing effect that prevents aggregation-based GNNs from easily amplifying the interactions between nodes and their neighbors, hence alleviating over-fitting and over-smoothing. Once training has finished, node masking can easily be inactivated by setting the node masking rate $p$ to 1.

\subsection{Node masking vs. Dropout techniques}
Recently, quite a few works have explored ideas around stochastically dropping nodes and edges within GNN layers using \textit{Dropout} \cite{dropout}. Rong et al. \cite{dropedge} proposed the concept of \textit{DropEdge} in spectral GNNs to alleviate the problem of over-fitting and over-smoothing. DropEdge randomly drops a subset of edges from the graph in every training epoch.
In the GAT paper \cite{velickovic2018graph}, the authors employed a similar technique by applying dropout to the attention coefficients within each GAT layer, which made the learning process more robust to over-fitting. However, neither of these techniques systematically addresses the problem of repetition in aggregation trees and neither of them facilitates caching. In fact, we show in our experiments that the GAT architecture performs much better when node masking is used instead of dropout on attention coefficients.


\section{Experiments}
\label{experiments}
To empirically verify the theory we have discussed up till now, we conduct over 200 experimental runs, covering both the aspects we highlighted, i.e., generalization and scalability.

\subsection{Datasets}
We work with three widely-used benchmark datasets for node classification: the \textit{Cora} and \textit{PubMed} citation networks \cite{datasets} and the social network graph of \textit{Reddit} posts \cite{hamilton2017inductive}. We pick these datasets since they are the ones used across several works that explore over-fitting and over-smoothing in GNNs \cite{wang2019improving,dropedge,smoothing,mad}.

\vspace{1mm}
\noindent
\textbf{Cora.} The \textit{Cora} citation network dataset consists of $2,708$ nodes and $5,429$ edges. Nodes denote scientific publications and edges denote the citation relationships amongst them. Note that the edges are undirected for the purpose of the dataset even though citations are not symmetric \cite{gcn,velickovic2018graph}. The publications are represented as binary bag-of-words vectors consisting of 1433 features each. Every node belongs to one of the seven classes, indicating the area of publication, e.g., \textit{Genetic Algorithms} or \textit{Reinforcement Learning}.

\vspace{1mm}
\noindent
\textbf{PubMed.} The \textit{PubMed diabetes} dataset consists of $19,717$ nodes and $44,338$ edges. Nodes denote scientific publications on diabetes and edges denote the citation relationships amongst them. As above, the edges are undirected for the purpose of the dataset. The publications are represented as \textit{TF-IDF} weighted bag-of-words feature vectors with 500 features each. Every node belongs to one of the three classes, indicating the type of diabetes that the publication is about, e.g., \textit{Diabetes Mellitus Type 1} or \textit{Diabetes Mellitus Type 2}.

\vspace{1mm}
\noindent
\textbf{Reddit.} The \textit{Reddit} posts graph consists of $232,965$ nodes denoting posts from $41$ different \textit{sub-reddits}. An undirected edge is present between two posts if the same user(s) commented on the two posts. Each post is represented by a $602$-dimensional feature vector formed by concatenating distributional semantic features and count-based features for title and comments.

\subsection{Models and configurations}
We experiment with both of the aggregation-based GNNs that we have discussed up till now, i.e., GAT and GIN. For the GIN architecture, we utilize the formulation specified by equation \ref{equation7}. If node masking is inactive, this formulation behaves exactly like the GIN-0 architecture from the original paper \cite{xu2018how} that was shown to have state of the art performance. For the GAT architecture, we use the original formulation \cite{velickovic2018graph} specified by equation \ref{equation2}. Additionally, we define a variant of GAT that we refer to as \textit{simple} GAT or \textit{SGAT}:

\begin{equation}
\label{equation9}
    h^{(k)}_v = \bigg\Vert_l\,\sigma\left(\,\sum_u^{v\,\cup\,\mathcal{N}_v}\frac{\delta^G_p(u)}{|v \cup \mathcal{N}_v|}\cdot W^{(k)}_lh^{(k-1)}_u\,\right)
\end{equation}

When node masking is inactive, this variant behaves exactly like the original GAT architecture except that the attention coefficients are now simply the inverse of a node's degree.

\subsection{Experimental settings}
We use \textit{PyTorch} \cite{pytorch} for modeling. For the GAT and SGAT architectures, we set the exact same hyper-parameters as in the original paper \cite{velickovic2018graph} for the \textit{Cora} dataset. We do not experiment with GAT and SGAT on the \textit{PubMed} and \textit{Reddit} datasets because that requires support for operations like \textit{sparse softmax} \cite{velickovic2018graph}, which are not stable in \textit{PyTorch} yet. Akin to GAT, we found a 2-layer GIN to be optimal for all the datasets. We set the maximum number of epochs to $1,000$ with an early stopping patience of $50$ epochs. We use the \textit{Adam} optimizer \cite{adam} to update the parameters. If node masking is active, then in every training epoch, we randomly sample a Bernoulli select function over the nodes in the train set. Node masking is always inactive outside training. We submit our code for further reference.

We experiment with both transductive and inductive learning paradigms. In the transductive setting, we make the entire graph from the dataset available at training time, i.e., all the nodes and edges. That said, the the loss for back-propagation is calculated using labels on the nodes in the train set only. In the inductive setting, we only make available at training time the graph $\mathcal{G}$ formed by nodes in the train set plus the edges amongst them. At validation and test times, we introduce the relevant nodes from the dataset into $\mathcal{G}$ along with the corresponding edges.

Unlike previous works, which test their hypotheses only on a single split of the datasets, we experiment with multiple splits of the three datasets in both the transductive and inductive settings. Specifically, for each dataset, we experiment with 10\%, 20\%, 25\%, 33\%, 50\%, 75\%, and 90\% of the nodes as the train set. In all the cases, the remaining data forms our test set except for a small part that we designate as the validation set for evaluating the early stopping criterion in the training phase. For every split, we perform stratified partitioning of the data to ensure similar class distribution.

Note that the metrics we present from our experiments are all in fact mean metrics over 10 trials with random initializations of the parameters. Unlike previous works that use micro F$_1$ and accuracy, we report macro F$_1$ since we strongly consider it to be a more appropriate metric as it provides an equal evaluation of effectiveness on all classes \cite{van2013macro}. That said, we noted that all the trends are consistent with micro F$_1$ and accuracy too.

\subsection{Generalization}
To show that node masking helps aggregation-based GNNs generalize better, we compare the performances yielded by the models when node masking is active versus when it is inactive. We denote the configurations where node masking is active by ``+NM'' and use a consistent rate of $p=0.5$. That said, $p$ is a hyper-parameter that can be adjusted for further gains. Appendix \ref{p-vary} explores the change in performance as $p$ and number of layers in the GNN vary.

\begin{table}[ht]
    \caption{Macro F$_1$ scores on the \textit{Cora} dataset in inductive and transductive settings for various sizes \textit{s} of the train set (in percentage). Numbers in bold are significantly better than their counterparts with $p$-value $<0.05$ under paired \textit{t-test}.}
    \label{cora-gat-results}
    \begin{center}
    \begin{small}
    \begin{sc}
    \begin{tabular}{c c | c c | c c}
        \toprule
        & \textit{s} & \textbf{gat} & \textbf{gat+nm} & \textbf{sgat} & \textbf{sgat+nm}\\
        \midrule
                           & 10 & 40.74 & \textbf{51.25} & 40.57 & \textbf{50.64}\\
                           & 20 & 68.40 & \textbf{74.65} & 68.50 & \textbf{74.82}\\
                           & 25 & 74.12 & \textbf{78.71} & 74.34 & \textbf{79.32}\\
        \textit{Inductive} & 33 & 80.44 & \textbf{82.56} & 80.30 & \textbf{82.74}\\
                           & 50 & 84.77 & \textbf{85.52} & 85.04 & \textbf{85.44}\\
                           & 75 & 87.00 & 87.03          & 87.16 & 87.29         \\
                           & 90 & 86.86 & 86.72          & 86.46 & \textbf{86.94}\\
        \midrule
                              & 10 & 82.01 & \textbf{82.77} & 82.17 & \textbf{83.46}\\
                              & 20 & 84.20 & \textbf{85.04} & 84.35 & \textbf{85.34}\\
                              & 25 & 84.95 & \textbf{85.38} & 85.12 & \textbf{85.88}\\
        \textit{Transductive} & 33 & 85.87 & \textbf{86.35} & 85.86 & \textbf{86.72}\\
                              & 50 & 86.61 & \textbf{87.35} & 87.32 & \textbf{87.61}\\
                              & 75 & 87.93 & 88.18          & 88.43 & 88.44         \\
                              & 90 & 87.25 & 87.36          & 87.24 & 87.22         \\
        \bottomrule
    \end{tabular}
    \end{sc}
    \end{small}
    \end{center}
\end{table}

Table \ref{cora-gat-results} shows the comparisons between GAT and GAT+NM and between SGAT and SGAT+NM on the \textit{Cora} dataset. GAT+NM and SGAT+NM outperform their counterparts across several splits in both the transductive and inductive settings. As expected, the gains are more pronounced in the inductive setting and amongst smaller sizes of the train set. Note that in GAT+NM and SGAT+NM, we do not use any dropout on the attention coefficients. So, the wins over GAT and SGAT clearly indicate that node masking is more effective than stochastically dropping edges because the latter does not systematically address the issue of repetition of nodes in aggregation trees. More specifically, dropout-based techniques simply make the aggregation trees sparser but do not strongly impede amplified interactions between nodes and their neighbors.

We further validate our reasoning by calculating the \textit{Mean Average Distance} (\textit{MAD}) values for the GAT and SGAT models with and without node masking. MAD is a metric recently proposed by Chen et al. \cite{mad} that quantifies the smoothing caused by a GNN by computing a cosine distance based scalar measure over the representations generated for all the nodes in a graph. The lower the MAD value, the higher the smoothing caused by the GNN. Table \ref{cora-gat-mad} presents the MAD values on the \textit{Cora} dataset. We see that node masking is significantly better at alleviating over-smoothing than the stochastic exclusion of edges. These results are in line with the observations of Wang et al. \cite{wang2019improving} who found the dropout on attention coefficients to not be a very effective tool against over-smoothing.

\begin{table}[ht]
    \caption{MAD values on the \textit{Cora} dataset in inductive and transductive settings with a train set of size 10\%. Numbers in bold are significantly better than their counterparts with $p$-value $<0.05$ under paired \textit{t-test}.}
    \label{cora-gat-mad}
    \begin{center}
    \begin{small}
    \begin{sc}
    \begin{tabular}{l | c | c}
        \toprule
        & \textit{Inductive} & \textit{Transductive}\\
        \midrule
        \textbf{gat} (-attn. dropout)      & 0.20 & 0.46\\
        \textbf{gat}                       & 0.23 & 0.52\\
        \textbf{gat + nm}                  & \textbf{0.29} & \textbf{0.57}\\
        \midrule
        \textbf{sgat} (-attn. dropout)     & 0.20 & 0.44\\
        \textbf{sgat}                      & 0.24 & 0.51\\
        \textbf{sgat + nm}                 & \textbf{0.27} & \textbf{0.56}\\
        \bottomrule
    \end{tabular}
    \end{sc}
    \end{small}
    \end{center}
\end{table}

Tables \ref{cora-gin-results}, \ref{pubmed-gin-results} and \ref{reddit-gin-results} compare GIN and GIN+NM on the \textit{Cora}, \textit{PubMed} and \textit{Reddit} datasets respectively. Again, node masking helps boost performance across almost all splits in both the transductive and inductive settings. Note that, due to the size of the \textit{Reddit} dataset, we could not run the experiments in transductive setting or inductive setting with train set $>50\%$ on our machines with \textit{NVIDIA P100} GPUs. Other works \cite{dropedge} have noted the same.

\begin{table}[ht]
    \caption{Macro F$_1$ scores on the \textit{Cora}, \textit{PubMed} and \textit{Reddit} datasets in inductive and transductive settings for various sizes \textit{s} of the train set (in percentage). Numbers in bold are significantly better than their counterparts with $p$-value $<0.05$ under paired \textit{t-test}.}
    \label{gin-results}
    \centering
    \subfigure[\textit{Cora}]{
        \begin{small}
        \begin{sc}
        \begin{tabular}{c | c c | c c}
            \toprule
            & \multicolumn{2}{c |}{\textit{Inductive}} & \multicolumn{2}{c}{\textit{Transductive}}\\[1.5mm]
            \textit{s} & \textbf{gin} & \textbf{gin+nm} & \textbf{gin} & \textbf{gin+nm}\\
            \midrule
            10 & 52.91 & \textbf{56.38} & 77.95 & \textbf{79.65}\\
            20 & 71.10 & \textbf{74.38} & 81.65 & \textbf{82.66}\\
            25 & 74.71 & \textbf{77.94} & 82.72 & \textbf{83.87}\\
            33 & 78.84 & \textbf{80.98} & 83.78 & \textbf{84.61}\\
            50 & 83.07 & \textbf{84.49} & 85.63 & \textbf{86.57}\\
            75 & 85.46 & \textbf{86.88} & 87.40 & \textbf{87.91}\\
            90 & 85.13 & \textbf{86.63} & 86.67 & \textbf{87.82}\\
            \bottomrule
        \end{tabular}
        \end{sc}
        \end{small}
        \label{cora-gin-results}
    }
    \subfigure[\textit{PubMed}]{
        \begin{small}
        \begin{sc}
        \begin{tabular}{c | c c | c c}
            \toprule
            & \multicolumn{2}{c |}{\textit{Inductive}} & \multicolumn{2}{c}{\textit{Transductive}}\\[1.5mm]
            \textit{s} & \textbf{gin} & \textbf{gin+nm} & \textbf{gin} & \textbf{gin+nm}\\
            \midrule
            10 & 77.51 & \textbf{78.31} & 83.65 & \textbf{84.34}\\
            20 & 82.11 & \textbf{82.86} & 84.61 & \textbf{85.20}\\
            25 & 82.95 & \textbf{83.94} & 85.03 & \textbf{85.47}\\
            33 & 83.78 & \textbf{84.46} & 85.35 & \textbf{85.84}\\
            50 & 85.25 & \textbf{85.79} & 85.89 & \textbf{86.42}\\
            75 & 85.99 & \textbf{86.53} & 86.38 & 86.55         \\
            90 & 86.42 & \textbf{86.80} & 86.43 & 86.61         \\
            \bottomrule
        \end{tabular}
        \end{sc}
        \end{small}
        \label{pubmed-gin-results}
    }
    \subfigure[\textit{Reddit}]{
        \begin{small}
        \begin{sc}
        \begin{tabular}{c | c c}
            \toprule
            & \multicolumn{2}{c}{\textit{Inductive}}\\[1.5mm]
            \textit{s} & \textbf{gin} & \textbf{gin+nm}\\
            \midrule
            10 & 78.12 & \textbf{87.09}\\
            20 & 83.38 & \textbf{87.73}\\
            25 & 86.76 & \textbf{88.74}\\
            33 & 87.57 & \textbf{88.94}\\
            50 & 88.93 & \textbf{90.97}\\
            \bottomrule
        \end{tabular}
        \end{sc}
        \end{small}
        \label{reddit-gin-results}
    }
\end{table}

We found the MAD values for GIN and GIN+NM to be almost identical, e.g., 0.65 for \textit{Cora} with a train set of size 10\%. This is intuitive given that GIN is a maximally powerful GNN that, in theory, can achieve injective mapping from nodes to representations \cite{xu2018how}. However, injectivity can lead to over-fitting \cite{DBLP:journals/corr/abs-1905-13422}, which node masking counters by stochastically augmenting the training graph to stress the relational inductive biases. We validate this by presenting some loss curves in appendix \ref{loss-curves} on the \textit{Cora} and \textit{PubMed} datasets with and without node masking under transductive and inductive settings. They highlight the regularizing effect of node masking.

\subsection{Scalability}
Tables \ref{cora-scale} and \ref{pubmed-scale} compare the performances of GIN on the \textit{Cora} and \textit{PubMed} datasets respectively with and without caching. For both the settings, each table also shows the number of unique nodes involved in the computations done by the model to generate predictions for the nodes introduced into the graph at the test time. Caching consistently reduces the number of unique nodes involved because only those nodes from the train set need to be involved that are 1-hop neighbors of the nodes introduced at the test time. The gains in efficiency from caching are more pronounced when the number of nodes present at the training time is significantly more than the number of nodes entering at the test time, a common scenario in the real-world. However, as conjectured before, the model performs worse with caching because the structure of aggregation trees it captures at the test time differs from the structure it learned to capture at the training time. That said, when node masking is used, the performance in fact exceeds that of GIN without caching.

\begin{table}[ht]
    \caption{Macro F$_1$ scores on the \textit{Cora} and \textit{PubMed} datasets in inductive setting with and without caching for various sizes \textit{s} of the train set (in percentage). F$_1$ scores in bold are significantly better than their counterparts with $p$-value $<0.05$ under paired \textit{t-test}.}
    \label{scale-results}
    \centering
    \subfigure[\textit{Cora}]{
        \begin{small}
        \begin{sc}
        \begin{tabular}{ c | c c | c c c}
            \toprule
            & \multicolumn{2}{c |}{\textit{GIN}} & \multicolumn{3}{c}{\textit{GIN + Caching}}\\[1.5mm]
            \textit{s} & \textbf{f$_1$} & \textbf{\#nodes} & \textbf{f$_1$} & \textbf{f$_1$} (\textbf{+nm}) & \textbf{\#nodes}\\
            \midrule
            10 & 52.91 & 2708 & 51.89 & \textbf{55.82} & \textbf{2698}\\
            20 & 71.10 & 2708 & 69.59 & \textbf{74.17} & \textbf{2672}\\
            25 & 74.71 & 2708 & 73.55 & \textbf{77.32} & \textbf{2644}\\
            33 & 78.84 & 2708 & 77.60 & \textbf{80.72} & \textbf{2605}\\
            50 & 83.07 & 2708 & 82.50 & \textbf{83.95} & \textbf{2426}\\
            75 & 85.46 & 2708 & 84.90 & \textbf{86.59} & \textbf{1781}\\
            90 & 85.13 & 2708 & 84.81 & \textbf{85.85} &  \textbf{919}\\
            \bottomrule
        \end{tabular}
        \end{sc}
        \end{small}
        \label{cora-scale}
    }
    \subfigure[\textit{PubMed}]{
        \begin{small}
        \begin{sc}
        \begin{tabular}{c | c c | c c c}
            \toprule
            & \multicolumn{2}{c |}{\textit{GIN}} & \multicolumn{3}{c}{\textit{GIN + Caching}}\\[1.5mm]
            \textit{s} & \textbf{f$_1$} & \textbf{\#nodes} & \textbf{f$_1$} & \textbf{f$_1$} (\textbf{+nm}) & \textbf{\#nodes}\\
            \midrule
            10 & 77.51 & 19717 & 76.25 & \textbf{78.33} & \textbf{19614}\\
            20 & 82.11 & 19717 & 81.47 & \textbf{82.75} & \textbf{19300}\\
            25 & 82.95 & 19717 & 82.24 & \textbf{83.77} & \textbf{19059}\\
            33 & 83.78 & 19717 & 82.83 & \textbf{84.15} & \textbf{18553}\\
            50 & 85.25 & 19717 & 84.27 & \textbf{85.34} & \textbf{16733}\\
            75 & 85.99 & 19717 & 85.46 & \textbf{86.01} & \textbf{11944}\\
            90 & 86.42 & 19717 & 85.87 & \textbf{86.61} &  \textbf{6682}\\
            \bottomrule
        \end{tabular}
        \end{sc}
        \end{small}
        \label{pubmed-scale}
    }
\end{table}

Here, one might seek a comparison with GraphSAGE \cite{hamilton2017inductive} which reduces the number of unique nodes involved in the computation of $h^k_v$ for a node $v$ by sub-sampling its $k$-hop neighborhood. Such sub-sampling, however, makes GraphSAGE lose a lot of information, rendering it worse than aggregation-based GNNs on these benchmark datasets \cite{shchur2018pitfalls}. We instead reduce the number of unique nodes involved in the computation carried out by aggregation-based GNNs by approximating $h^k_v$ using the cached representations of 1-hop neighbors of $v$. This approximation does not degrade the performance when node masking is employed.


\section{Conclusion}
\label{conclusion}
In this paper, we introduced \textit{node masking}, a novel technique that significantly improves the performance of state of the art graph neural networks (GNNs). We first discussed some theoretical tools to better visualize the operations performed by spatial aggregation-based GNNs. Using these tools, we highlighted the issues that limit the ability of such GNNs to generalize and scale. Finally, we empirically demonstrated the effectiveness of node masking in enhancing the performance of aggregation-based GNNs on three widely-used benchmark datasets for node classification, the \textit{Cora} and \textit{PubMed} citation network and the social network graph of \textit{Reddit} posts. The observed trends also hold for the \textit{CiteSeer} dataset but we omitted it for brevity. Previous works utilizing these datasets had tested their hypotheses only on a single split of the datasets. We instead showed the efficacy of node masking on a range of splits under both the transductive setting as well as the inductive setting, hence laying down strong benchmarks for future research.

\bibliographystyle{ACM-Reference-Format}
\bibliography{sample-base}

\clearpage
\appendix
\section{Varying the node masking rate}
\label{p-vary}
The analyses presented in this section are with the \textit{Cora} dataset but the observed trends hold across all the datasets.

\begin{figure}[ht]
    \centering
    \subfigure[GIN+NM in inductive setting]{
        \includegraphics[width=8cm]{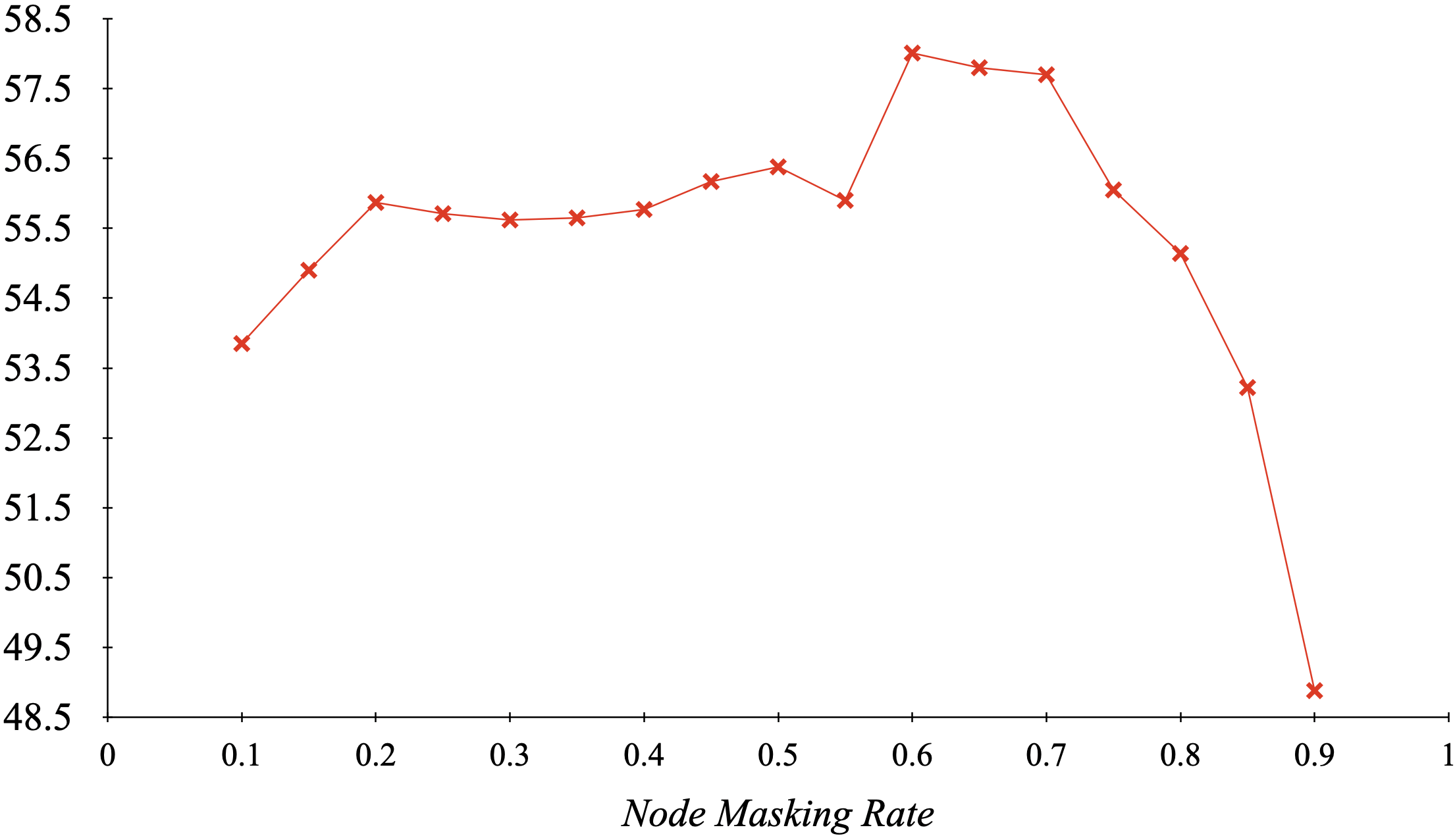}
    }
    \subfigure[GIN+NM in transductive setting]{
        \includegraphics[width=8cm]{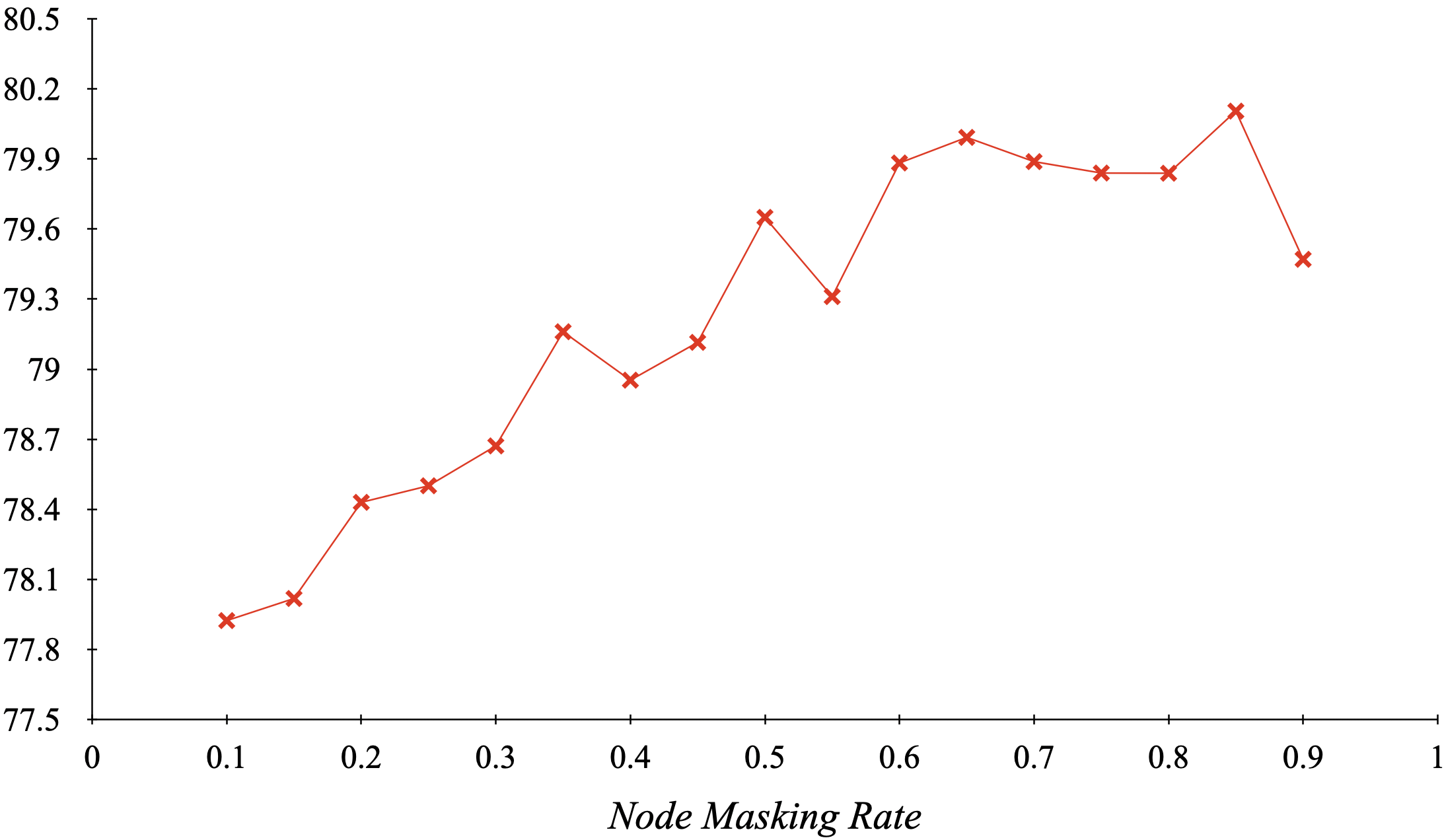}
    }
    \caption{Results on the \textit{Cora} dataset with GIN+NM under both the inductive and transductive settings for various node masking rates $p\in[0.1, 0.9]$. The train set size is 10\%. \textit{Y}-axis denotes macro F$_1$. We see that the best performance is not necessarily with $p=0.5$.}
    \label{varying-p-gin}
\end{figure}

\begin{figure}[ht]
    \centering
    \subfigure[GAT+NM in inductive setting]{
        \includegraphics[width=8cm]{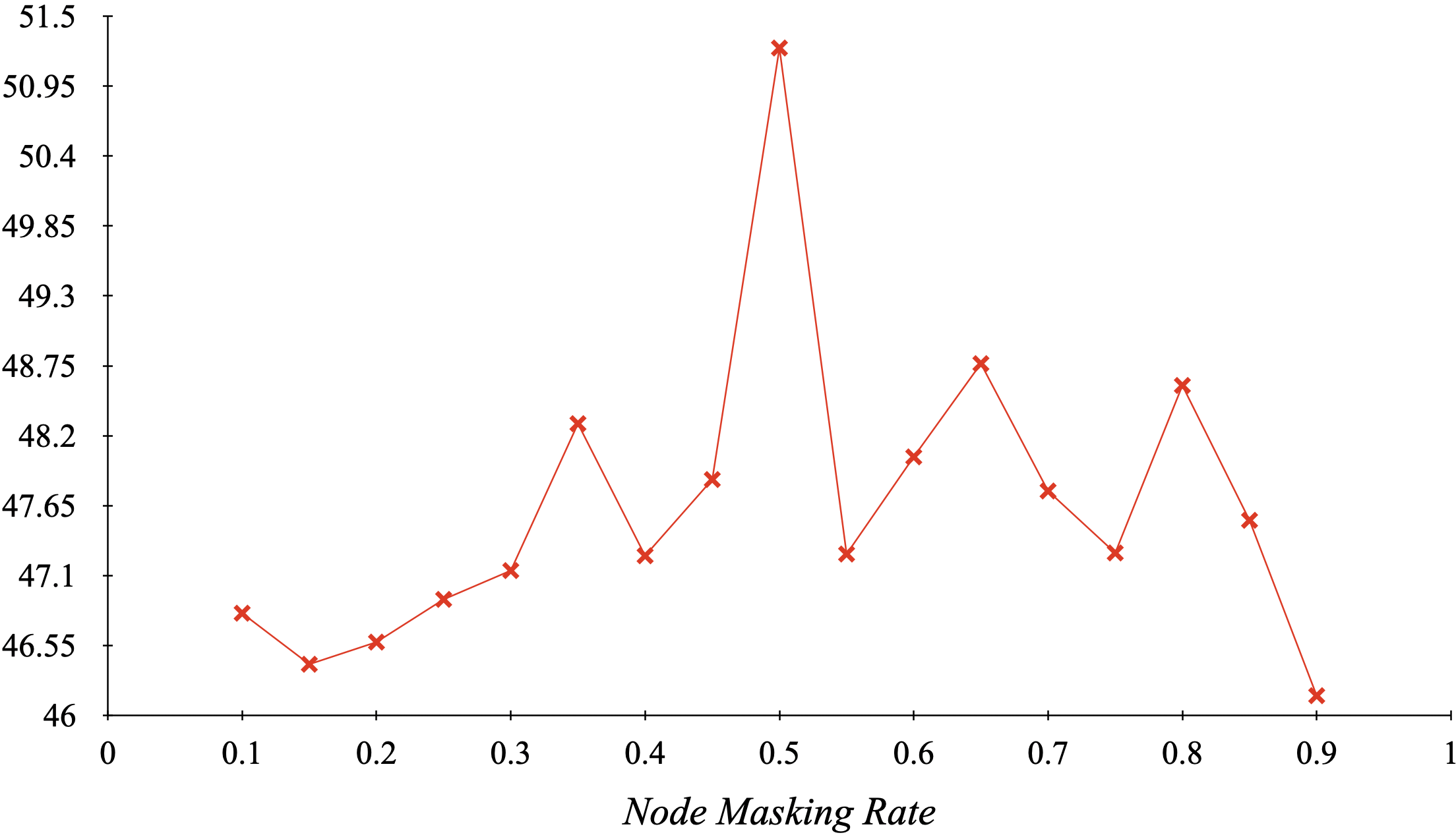}
    }
    \qquad\qquad
    \subfigure[SGAT+NM in inductive setting]{
        \includegraphics[width=8cm]{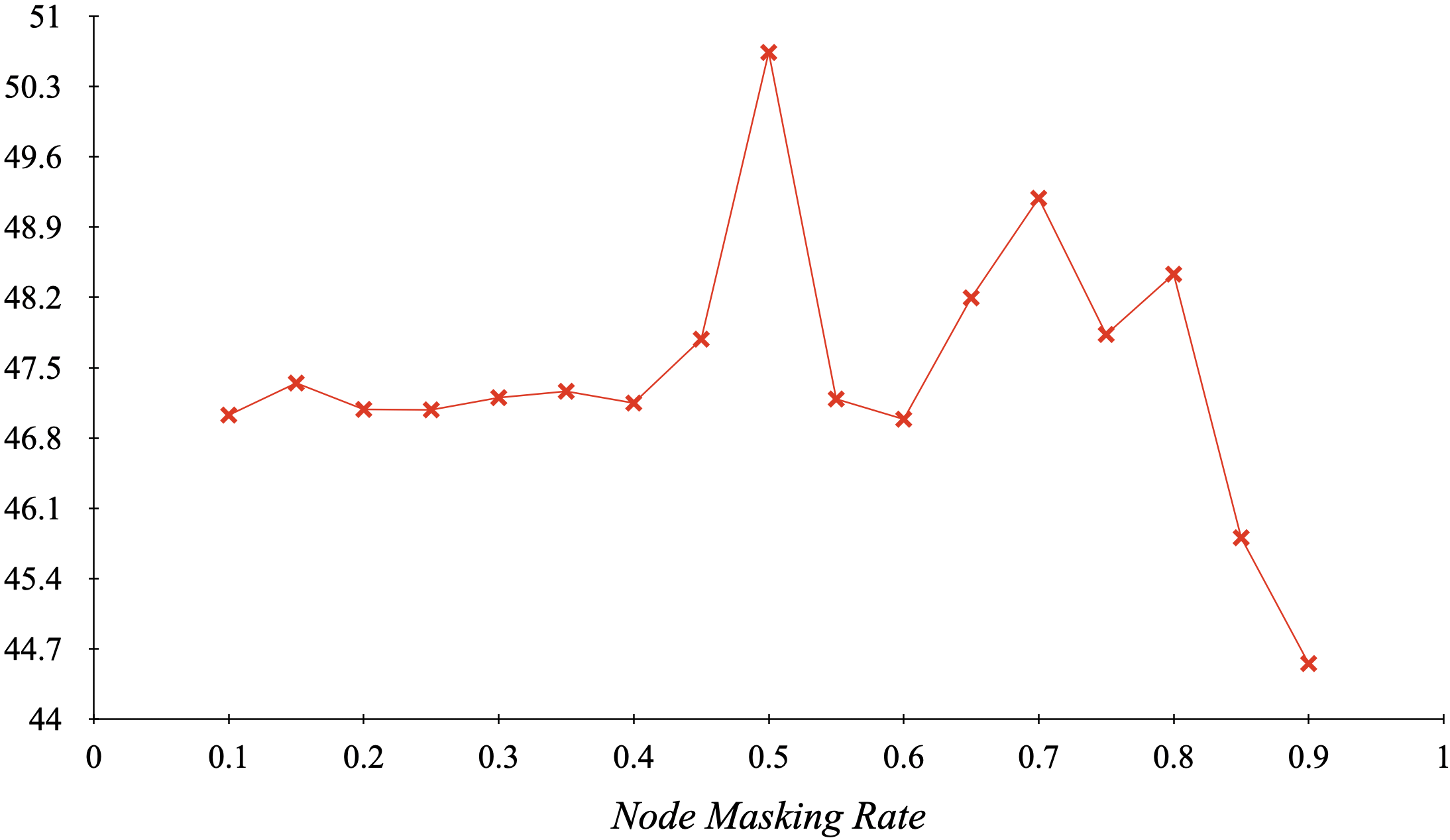}
    }
    \caption{Results on the \textit{Cora} dataset with GAT+NM and SGAT+NM under inductive setting for various node masking rates $p\in[0.1, 0.9]$. The train set size is 10\%. \textit{Y}-axis denotes macro F$_1$. We see that the best performance is with $p=0.5$.\vspace{15mm}}
    \label{varying-p-gat}
\end{figure}

Figures \ref{varying-p-gin} and \ref{varying-p-gat} explore the change in performance of GIN+NM, GAT+NM, and SGAT+NM as the node masking rate $p$ increases from 0.1 up to 0.9. We conduct the experiments on the \textit{Cora} dataset with a train set of size 10\%.

As seen in figure \ref{varying-p-gin}, GIN+NM achieves the best performance under both the transductive setting as well as the inductive setting at $p > 0.5$. This suggests that fine-tuning the node masking rate $p$ can yield even higher gains than we report in the paper. On the hand, figure \ref{varying-p-gat} suggests that GAT+NM and SGAT+NM achieve the best performance at $p=0.5$ only.

\begin{figure}[ht]
    \centering
    \includegraphics[width=8cm]{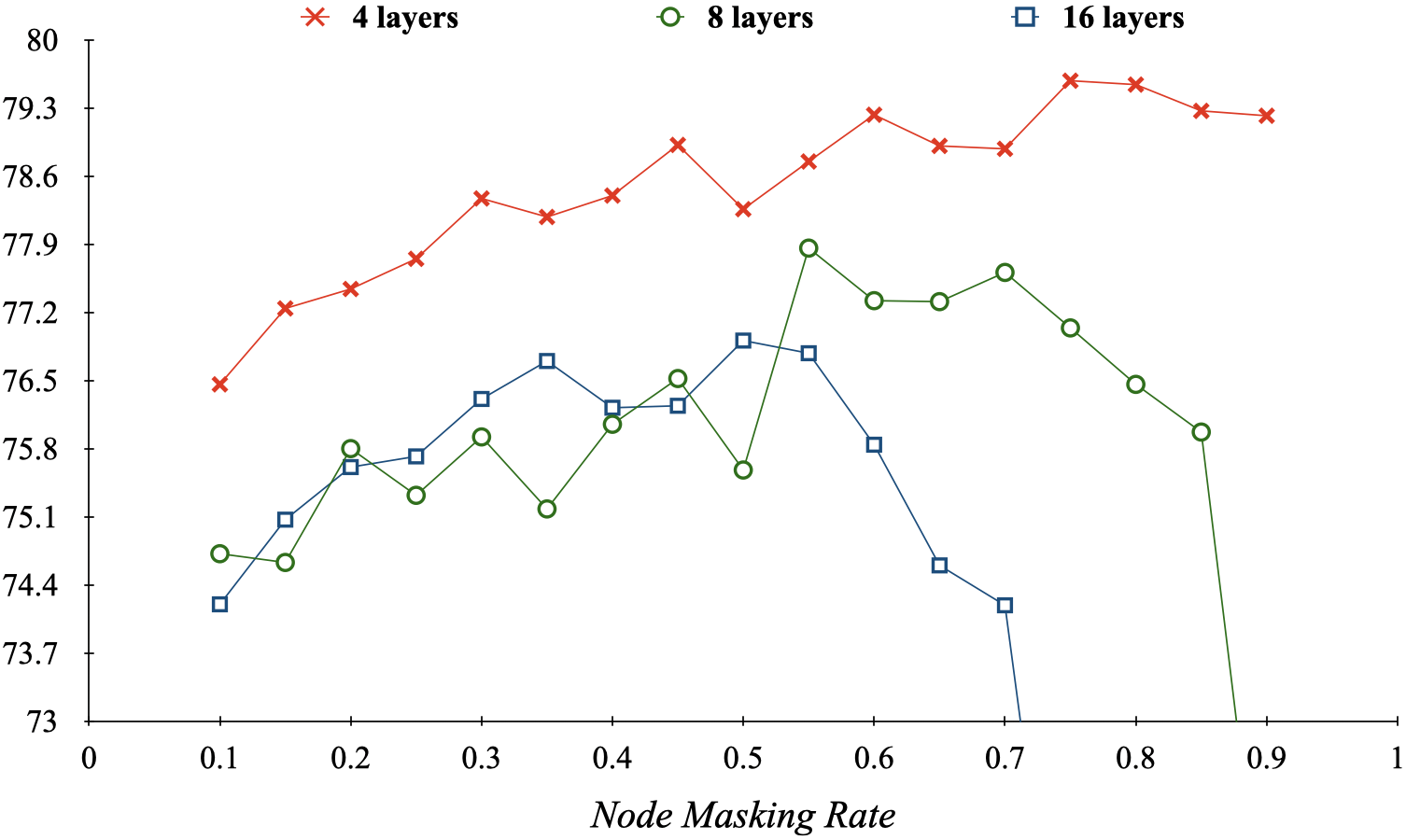}
    \caption{Results on the \textit{Cora} dataset with GIN+NM under the transductive setting for various node masking rates $p\in[0.1, 0.9]$ and depths of the model. The train set size is 10\%. \textit{Y}-axis denotes macro F$_1$. We see that the $p$ at which the optimal performance is achieved decreases with increasing depth.}
    \label{varying-layers-p-gin}
\end{figure}

Figure \ref{varying-layers-p-gin} further explores how the performance of GIN+NM changes with varying node masking rates $p$ when the number of GNN layers, i.e., the depth of the model, is increased from 2 to 4, 8 or 16. We note that as the depth increases, the $p$ at which the optimal performance is achieved decreases. This is intuitive given that more layers means more parameters, which in turn means denser aggregation trees are required for optimal training.

\clearpage
\section{Analysis of loss curves}
\label{loss-curves}
\begin{figure}[b]
    \centering
    \subfigure[Loss curves on nodes in the train set]{
        \includegraphics[width=8cm]{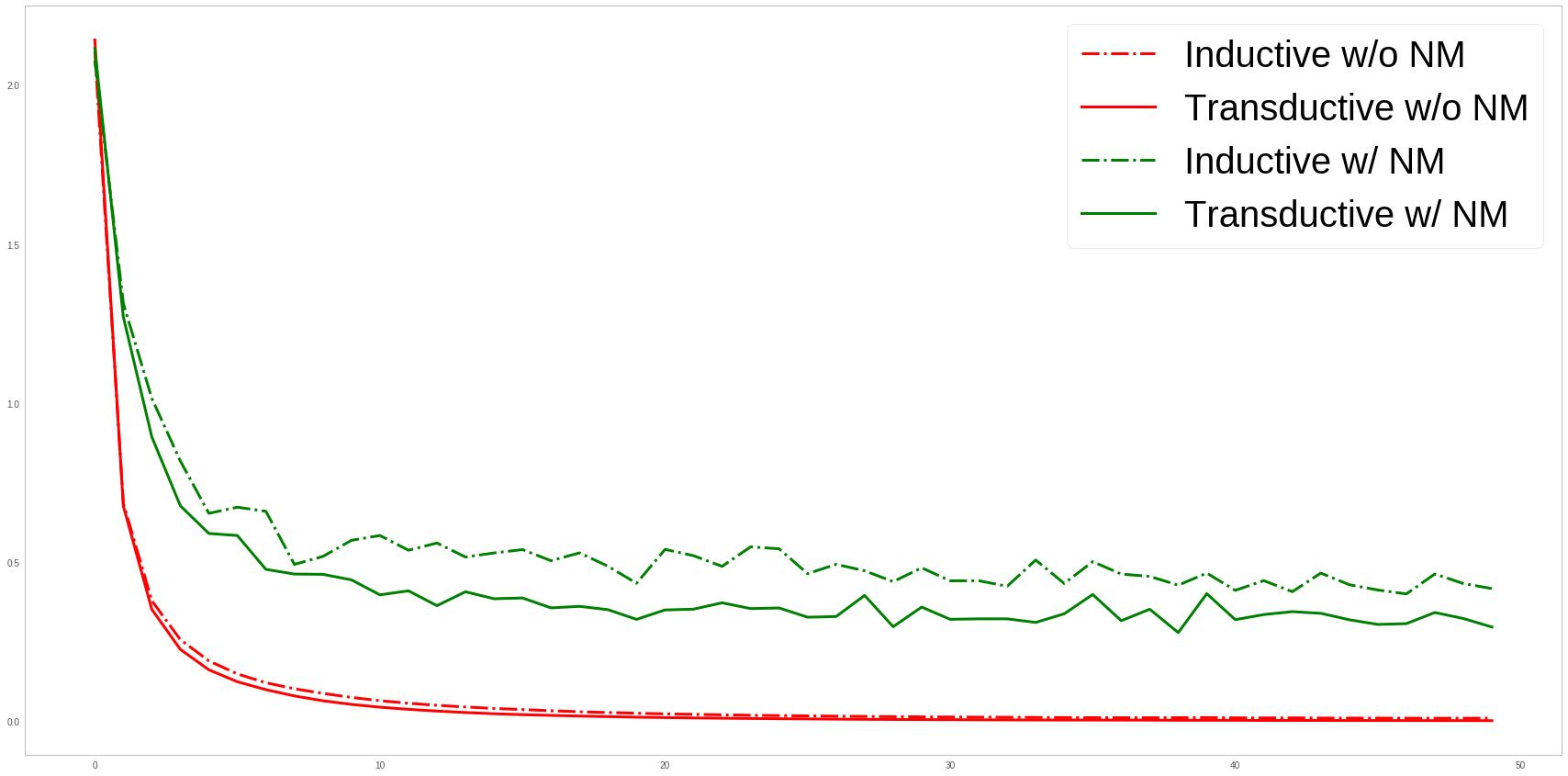}
    }
    \subfigure[Loss curves on nodes in the validation set]{
        \includegraphics[width=8cm]{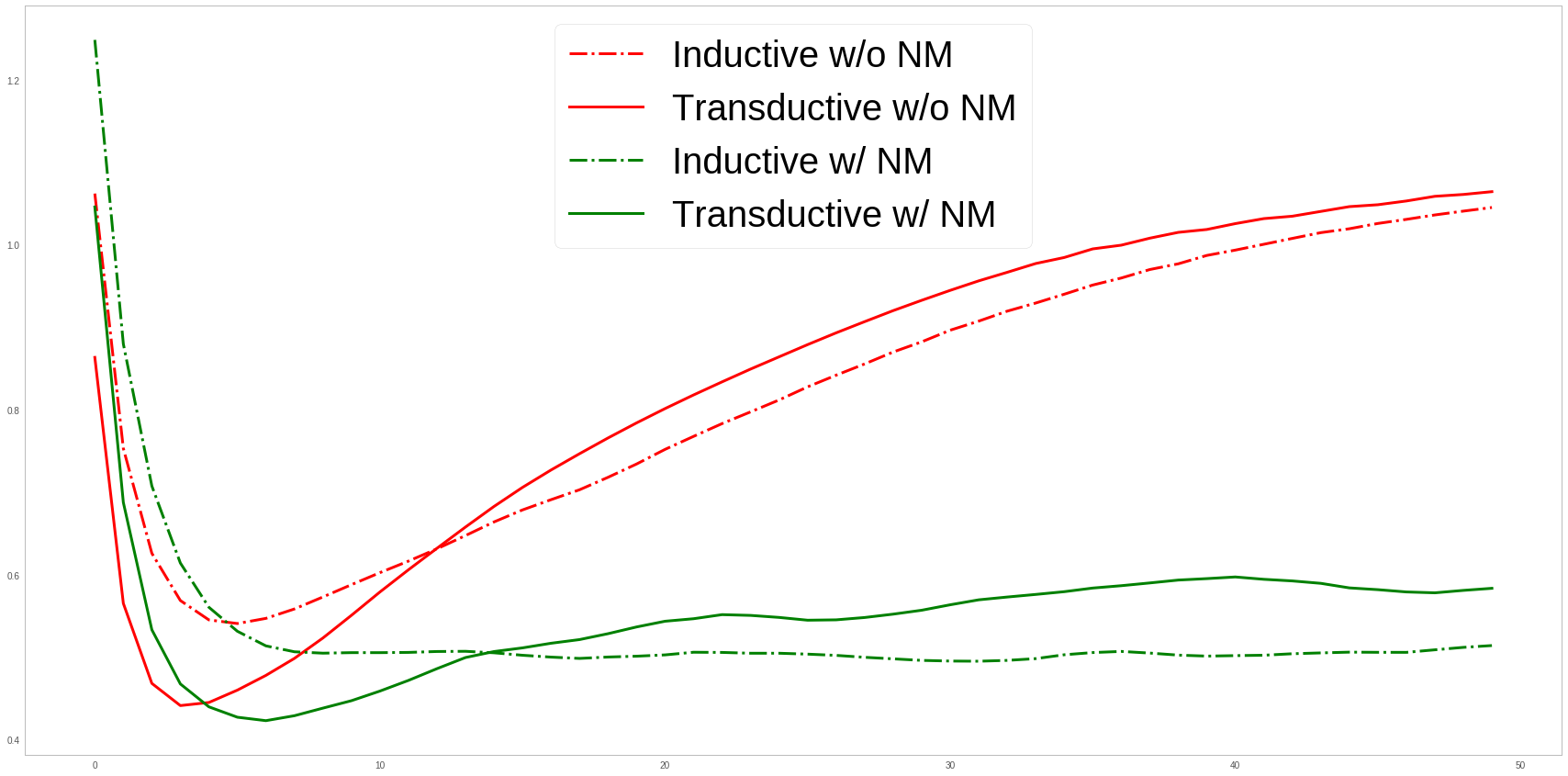}
    }
    \caption{Loss curves on the \textit{Cora} dataset for the GIN model with (w/ NM) and without (w/o NM) node masking.}
    \label{cora-gin-loss-curves}
\end{figure}

\begin{figure}[b]
    \centering
    \subfigure[Loss curves on nodes in the train set]{
        \includegraphics[width=8cm]{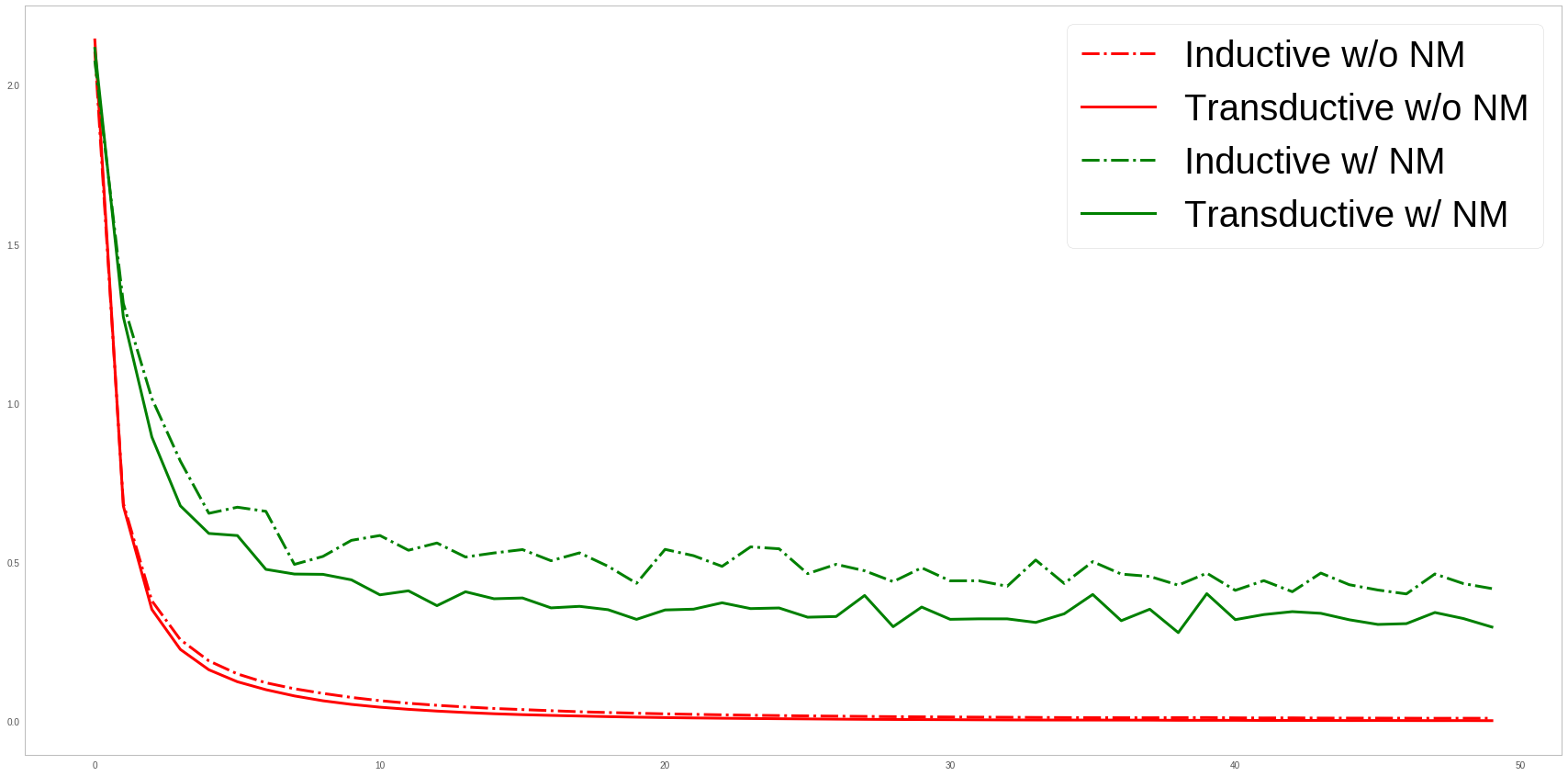}
    }
    \subfigure[Loss curves on nodes in the validation set]{
        \includegraphics[width=8cm]{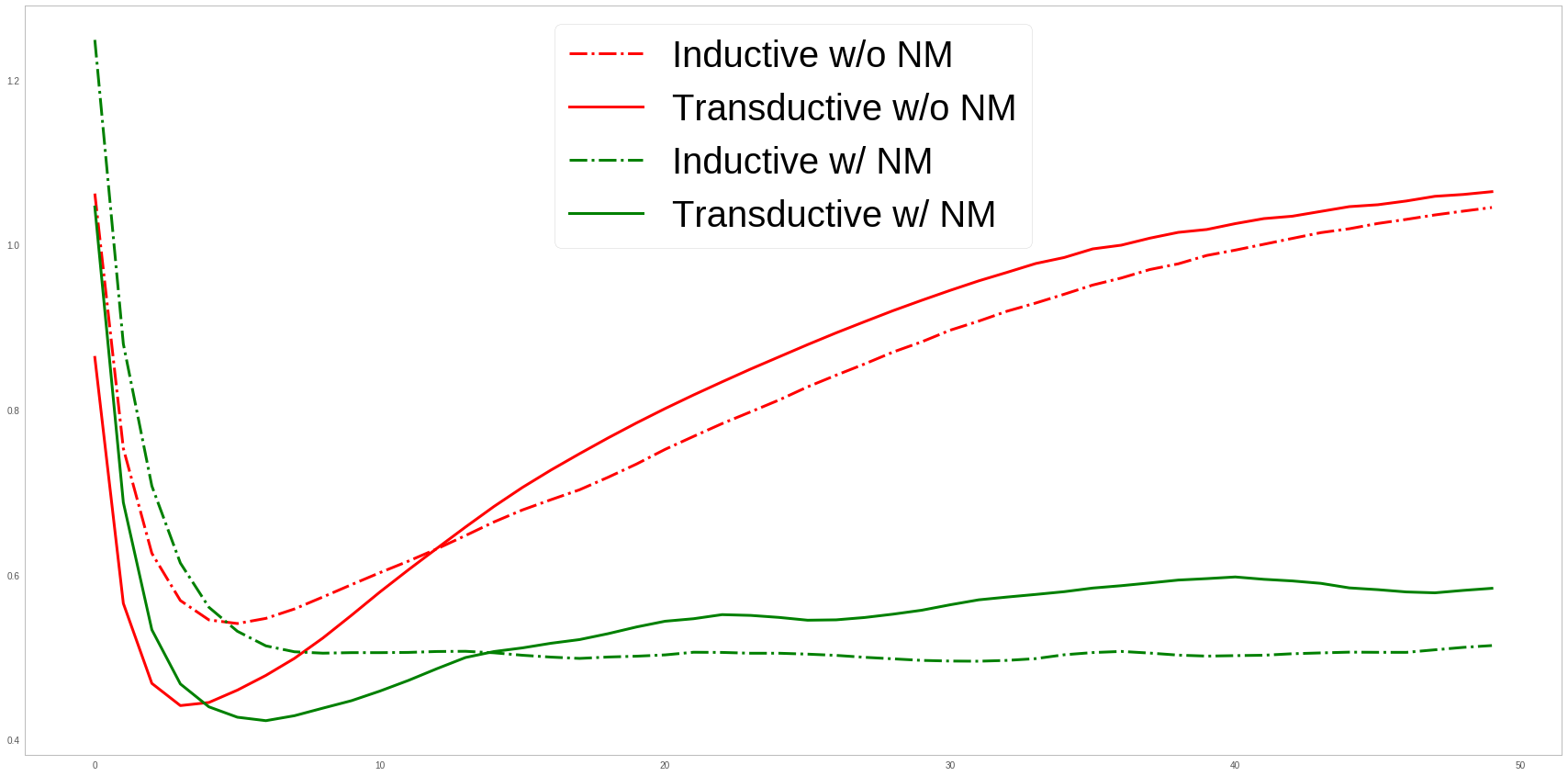}
    }
    \caption{Loss curves on the \textit{PubMed} dataset for the GIN model with (w/ NM) and without (w/o NM) node masking.}
    \label{pubmed-loss-curves}
\end{figure}

Figures \ref{cora-gin-loss-curves} and \ref{pubmed-loss-curves} present some loss curves from the training and validation phases for the GIN model with (green) and without (red) node masking under both the transductive and inductive settings.

As can be noted, node masking leads to higher training losses in both the transductive and inductive settings. This is typical of a regularization effect. Additionally, we also observe that the validation losses go up sharply after a point when node masking is not used, clearly indicating that the model has over-fit. With node masking, such an over-fitting phenomenon is neither observed in the transductive setting nor in the inductive setting. Therefore, the model is able to achieve a lower validation loss meaning that it generalizes better. This is typical of a higher inductive bias effect.

\end{document}